\documentclass[10pt,journal,compsoc]{IEEEtran}

\ifCLASSOPTIONcompsoc
  \usepackage[nocompress]{cite}
\else
  \usepackage{cite}
\fi
\ifCLASSINFOpdf
\else
\fi

\usepackage{enumitem}
\usepackage{amsfonts}
\usepackage{amsmath}
\usepackage{multirow}
\usepackage{graphicx}
\usepackage{algorithm}
\usepackage{algorithmicx}
\usepackage{algpseudocode}
\usepackage{booktabs}
\usepackage{bm}
\usepackage{color}
\usepackage{url}
\usepackage[table,dvipsnames]{xcolor}
\usepackage{colortbl}

\newcommand{\etal}{\emph{et al.}}

\def\btheta{{\boldsymbol \theta}}
\def\bphi{{\boldsymbol \phi}}

\def\tphi{\tilde{\boldsymbol{\phi}}}
\def\hphi{\hat{\boldsymbol{\phi}}}

\newcommand{\red}[1] {\textcolor[rgb]{1.0,0.0,0.0}{{#1}}}

\newcommand{\gain}[1] {\color{OliveGreen}{{#1}}}

\begin{document}
%

\title{Rethinking Autoregressive Models for Lossless Image Compression via Hierarchical Parallelism and Progressive Adaptation}
\author{Daxin Li, Yuanchao Bai,~\IEEEmembership{Member,~IEEE,} Kai Wang, Wenbo Zhao,~\IEEEmembership{Member,~IEEE,},\\ Junjun Jiang,~\IEEEmembership{Senior Member,~IEEE,} Xianming Liu,~\IEEEmembership{Member,~IEEE}
\IEEEcompsocitemizethanks{
\IEEEcompsocthanksitem Daxin Li, Yuanchao Bai,  Kai Wang, Wenbo Zhao, Junjun Jiang and Xianming Liu are with the School of Computer Science and Technology, Harbin Institute of Technology, Harbin, 150001, China. E-mail: \{hahalidaxin, cswangkai\}@stu.hit.edu.cn, \{yuanchao.bai, wbzhao, jiangjunjun, csxm\}@hit.edu.cn. \protect \\
}
\thanks{(Corresponding author: Yuanchao Bai)}
}

%
%

\markboth{Journal of \LaTeX\ Class Files,~Vol.~14, No.~8, August~2015}%
{Shell \MakeLowercase{\textit{et al.}}: Bare Advanced Demo of IEEEtran.cls for IEEE Computer Society Journals}
%



\IEEEtitleabstractindextext{%
\begin{abstract}
Autoregressive (AR) models have long set the theoretical performance benchmark for learned lossless image compression, but are often dismissed as impractical due to their prohibitive computational cost. This work re-thinks this paradigm, introducing a novel framework built on the twin principles of hierarchical parallelism and progressive adaptation that re-establishes pure autoregression as a top-performing and practical solution.
Our hierarchical parallel approach is embodied in the Hierarchical Parallel Autoregressive ConvNet (HPAC), an ultra-lightweight pre-trained model that uses a hierarchical factorized structure and a content-aware convolutional gating mechanism to efficiently capture spatial dependencies. To make this architecture practical, we introduce two key optimizations: Cache-then-Select Inference (CSI), which dramatically accelerates coding by eliminating redundant computations, and Adaptive Focus Coding (AFC), which efficiently extends the framework to high bit-depth images by managing their large alphabets.
Building on this lightweight and efficient foundation, which inherently reduces the computational burden of the forward and backward passes, our progressive adaptation strategy is realized by Spatially-Aware Rate-Guided Progressive Fine-tuning (SARP-FT). This efficient instance-level strategy fine-tunes the model for each test image by optimizing low-rank adapters on progressively larger, spatially-continuous regions selected based on their estimated information density.
Extensive experiments on diverse datasets, including natural, satellite, and medical images, validate that our method achieves new state-of-the-art compression performance. Notably, our approach sets a new benchmark in learned lossless compression, demonstrating that a carefully designed AR framework can offer significant gains over existing methods while maintaining a remarkably small parameter count and competitive coding speeds.
\end{abstract}

}

\maketitle

\IEEEdisplaynontitleabstractindextext

%
\IEEEpeerreviewmaketitle

%

\section{Introduction}

\IEEEPARstart{L}{ossless} image compression is a fundamental technology for digital imaging, enabling the efficient storage and transmission of image data while guaranteeing perfect reconstruction. Its significance is especially pronounced in domains where fidelity and precision are non-negotiable, such as medical imaging, satellite remote sensing, and scientific archiving. In these scenarios, even the slightest loss or alteration of information can have serious consequences, making lossless compression indispensable for ensuring data integrity and authenticity.

With the rapid proliferation of high-resolution and high bit-depth images, there is a growing demand for universal lossless compression methods that can robustly handle diverse image types. Traditional codecs like PNG, JPEG-LS~\cite{weinberger2000loco}, and JPEG-XL~\cite{alakuijala2019jpeg} rely on hand-crafted heuristics and, while effective, often fall short of the compression performance achieved by modern learned methods.

Recent advances in deep learning have introduced powerful generative probabilistic models for compression. While this paradigm shift has revolutionized lossy compression through end-to-end optimized VAE-based frameworks~\cite{hu2021learning, duan2023qarv,yu2024robust, li2024groupedmixer}, the application to lossless compression has seen parallel developments.
Among these, autoregressive (AR) models~\cite{van2016conditional, mentzer2019practical} have consistently set the theoretical performance benchmark, capturing intricate data distributions to achieve top-tier compression ratios. However, this performance has come at a cost. The strict causal dependency of pure AR models leads to notoriously slow, sequential inference. Consequently, the field has largely shifted towards alternative architectures, such as VAE-based models~\cite{mentzer2019practical, kingma2019bit}, flow-based models~\cite{zhang2021iflow, hoogeboom2019integer}, and other parallel-friendly hybrid methods~\cite{rhee2022lc,bai2024deep,Zhang2024ArIBBPS}. These approaches achieve practical coding speeds by trading away a degree of compression performance, often at the cost of a heavy parameter count.

This has left two fundamental challenges unresolved: 1) Is it possible to design an AR-based model that retains its state-of-the-art (SOTA) performance while being lightweight and efficient? 2) How can such a pre-trained model be efficiently adapted to overcome the generalization gap and achieve universal compression on unseen, out-of-domain images?

This paper re-thinks the conventional wisdom by answering 'yes' to both questions. We challenge the notion that autoregressive models are inherently impractical. We introduce a novel framework built on the twin principles of hierarchical parallelism and progressive adaptation, demonstrating that a carefully designed AR model can be simultaneously fast, compact, and efficiently adapted to individual images.

Our approach is built upon two core innovations corresponding to these principles. First, to achieve hierarchical parallelism, we introduce the Hierarchical Parallel Autoregressive ConvNet (HPAC), an ultra-lightweight pre-trained model. HPAC employs a hierarchical factorized structure and a content-aware convolutional gating mechanism to efficiently capture both intra- and inter-patch dependencies. To tackle the primary inference bottleneck, we introduce Cache-then-Select Inference (CSI), an optimized scheme that dramatically accelerates coding by eliminating redundant computations in masked convolutions. Furthermore, we extend the HPAC framework to high bit-depth images using Adaptive Focus Coding (AFC), a dynamic PMF truncation and coding strategy.

Crucially, this lightweight and efficient architecture does more than solve the inference bottleneck. By inherently reducing the computational burden of both forward and backward passes, it serves as the essential foundation that enables rapid test-time adaptation—our second challenge.

Second, building on this efficient foundation to enable progressive adaptation, we introduce a strategy inspired by the Minimum Description Length (MDL) principle. Our method, Spatially-Aware Rate-Guided Progressive Fine-tuning (SARP-FT), fine-tunes the pre-trained HPAC for each test image using the Parameter-Efficient Transfer Learning (PETL) technique of low-rank adaptation. SARP-FT progressively updates the model on spatially-continuous regions, prioritized by their estimated information density. This allows the model to converge quickly on image-specific details, with the minimal adapted parameters encoded as a compact prompt alongside the image.

Our contributions are encapsulated as follows:
\begin{itemize}
    \item We present the Hierarchical Parallel Autoregressive ConvNet (HPAC), an ultra-lightweight architecture built on a hierarchical parallel structure. To make it practical, we introduce Cache-then-Select Inference (CSI) for acceleration and Adaptive Focus Coding (AFC) to efficiently handle high bit-depth images.
    \item We introduce Spatially-Aware Rate-Guided Progressive Fine-tuning (SARP-FT), a novel strategy unifying the MDL principle and PETL. It enables rapid and efficient instance-specific adaptation by progressively fine-tuning low-rank adapters on spatially-contiguous, high-information-density regions.
    \item Through extensive experiments, we demonstrate that our combined framework achieves new state-of-the-art performance across diverse datasets, significantly outperforming existing methods while maintaining a remarkably small parameter count and competitive coding speeds.
\end{itemize}

This manuscript is a substantial extension of our conference paper~\cite{li2024callic}. New contributions include: 1) A novel hierarchical parallel framework in HPAC that evolves beyond the conference work's intra-patch focus to exploit both intra- and inter-patch dependencies; 2) The Cache-then-Select Inference (CSI) scheme, an optimization that further reduces redundant computations in masked convolutions for enhanced efficiency; 3) The Adaptive Focus Coding (AFC) extension to efficiently manage large alphabets for high bit-depth images; 4) A revised progressive adaptation strategy (SARP-FT), which is now designed to select spatially-contiguous regions to align with the new hierarchical model's inter-patch dependency structure; and 5) Comprehensive additional experiments validating the method's superior performance.

The remainder of this paper is structured as follows. Section \ref{sec:related_work} reviews related work. Section \ref{sec:foundations} establishes the theoretical foundations for our approach. Section \ref{sec:method} details the implementation of our two-part framework: Section \ref{ssec:overall_architecture} introduces the HPAC model, CSI, and high bit-depth extensions (AFC), while Section \ref{ssec:adaptation} describes the MDL-principled progressive adaptation strategy (SARP-FT). Section \ref{sec:experiments} presents experimental results. Finally, Section \ref{sec:conclusion} concludes the paper.

\section{Related Work \label{sec:related_work}}

\subsection{Autoregressive Models for Lossless Compression}
Autoregressive (AR) models factorize the joint pixel probability into a product of conditional probabilities via the chain rule. PixelCNN~\cite{van2016conditional} pioneered this deep learning approach, using masked convolutions to model these conditional distributions. PixelCNN++~\cite{salimans2016pixelcnn++} later improved upon this foundation, notably by introducing a discretized logistic mixture model for the outputs.

Due to their expressive power, AR models have consistently set the theoretical performance benchmark for lossless compression~\cite{van2016conditional,mentzer2019practical, child2019generating}. However, their strict sequential nature results in prohibitively slow inference, leading them to be widely considered impractical for real-world applications. Subsequent work has largely focused on mitigating this stark performance-efficiency trade-off, primarily by creating hybrid models

One popular direction is to apply a lightweight or less complex AR model to the residuals of a faster, parallel predictor. For instance, Bai \etal~\cite{bai2021learning} proposed a framework combining fast lossy prediction (using a Variational Autoencoder or VAE) with a shallow, serial AR model to encode the residual; this approach was later parallelized for faster processing~\cite{bai2024deep}. Rhee \etal~\cite{rhee2022lc} also extended this residual-based framework by modeling autoregressive dependencies across spatial, illumination, and color components. Similarly, PILC~\cite{kang2022pilc} achieves practical speeds by using a VAE to predict residuals, which are then modeled by a shallow, diagonal- and column-parallel AR model.

Another approach involves integrating AR priors with hierarchical latent variables. Split Hierarchical VAE~\cite{ryder2022split} combines a layer-wise 4-step autoregressive prior with a hierarchical VAE structure to improve compression and solve issues related to bits-back coding. Extending this, ArIB-BPS~\cite{Zhang2024ArIBBPS} combines bit plane slicing with VAEs and a 4-step autoregressive component. 

Although these hybrid models achieve better coding speeds than conventional serial AR models, this gain often comes at the cost of compromised compression performance—stemming from weaker autoregressive modeling or restrictive architectural choices—and a high parameter count, often introduced by the latent variable components. This motivates our work to rethink the pure autoregressive model from the ground up, questioning if its long-standing sequential bottleneck can be overcome with carefully designed hierarchical parallelism.

\subsection{Test-Time Adaptation in Image Compression}
A separate but equally critical challenge for any pre-trained compression model is the amortization gap: they are trained to generalize across a large dataset, which limits their ability to model the unique statistics of a single test image, especially one from an out-of-domain (OOD) source. Test-time adaptation is a powerful solution, but fine-tuning an entire large model for every image is computationally infeasible.

This is where Parameter-Efficient Transfer Learning (PETL) becomes critical. PETL enables adaptation of pre-trained models by fine-tuning only a small fraction of parameters~\cite{ding2023parameter}. PETL methods generally fall into three categories:
\begin{itemize}
\item \textit{Adapter Modules}~\cite{houlsby2019parameter} insert lightweight trainable layers (e.g., bottleneck layers) between frozen backbone layers, enabling efficient adaptation while keeping most parameters fixed~\cite{rebuffi2017learning,chen2022adaptformer}.
\item \textit{Prompt Tuning} optimizes learnable prompt vectors that are appended to model inputs or intermediate features~\cite{li2021prefix,lester2021power}. These prompts are learned directly for downstream tasks while keeping the backbone frozen.
\item \textit{Low-Rank Adaptation} (LoRA)~\cite{hu2021lora} has become a prominent technique. It freezes the original weights and injects trainable, low-rank decomposition matrices into model layers, significantly reducing trainable parameters while maintaining strong adaptation capacity. Recent variants like DoRA~\cite{liu2024dora} and QLoRA~\cite{dettmers2023qlora} further improve efficiency.
\end{itemize}

In image compression, PETL techniques have been explored mainly in lossy frameworks, for example, using adapter modules to improve generalization~\cite{tsubota2023universal,shen2023dec}. Its application to lossless compression is less mature. In lossless compression, Zhang \etal~\cite{zhang2025fitted} concurrently proposed FNLIC, which combines pre-trained autoregressive and overfitted VAE models. However, their approach is impractically slow, requiring over an hour per image. This highlights a critical need not just for an adaptive method, but for an efficiently adaptive one. Our framework directly addresses this, achieving superior compression while being significantly faster (e.g., 8.4s per image, $\sim$700$\times$ faster than FNLIC).

This review highlights two persistent gaps in lossless compression: 1) \textbf{The Performance-Efficiency Gap}, where SOTA performance (from AR models) is sacrificed for practical speeds (from hybrids and alternatives). 2) \textbf{The Universality-Adaptation Gap}, where pre-trained models fail on OOD data, and existing adaptation methods are impractically slow. Our work confronts both challenges simultaneously. We address the first gap by rethinking the AR model itself through hierarchical parallelism and its efficient implementing. We address the second gap with an efficient progressive adaptation strategy.

\begin{figure*}[t]
 \centering
 \includegraphics[width=1.\linewidth]{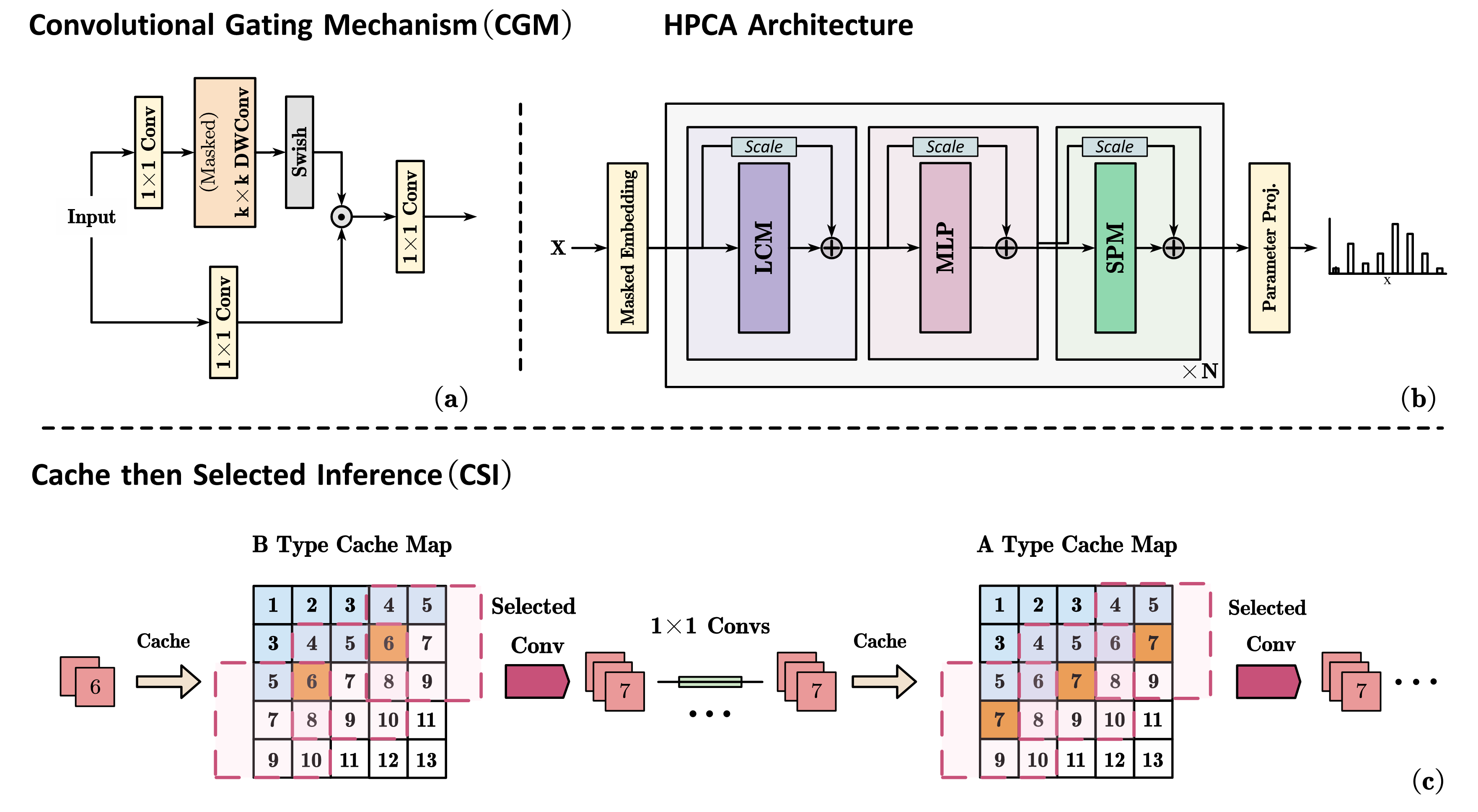}
 \caption{Overview of the proposed architectures and mechanisms in HPAC: (a) The content-adaptive Convolutional Gating Mechanism (CGM) simplifies attention. (b) Building on this, the Hierarchical Parallel Autoregressive ConvNet (HPAC) consists of Local Context Modulator (LCM), MLP, and Spatial Propagation Module (SPM) blocks. (c) To accelerate coding, Cache-then-Select Inference (CSI) caches activations and performs efficient selective computation only on causally relevant features.
 }
\label{fig:network}
\end{figure*}


\begin{figure}[t]
 \centering
 \includegraphics[width=\linewidth]{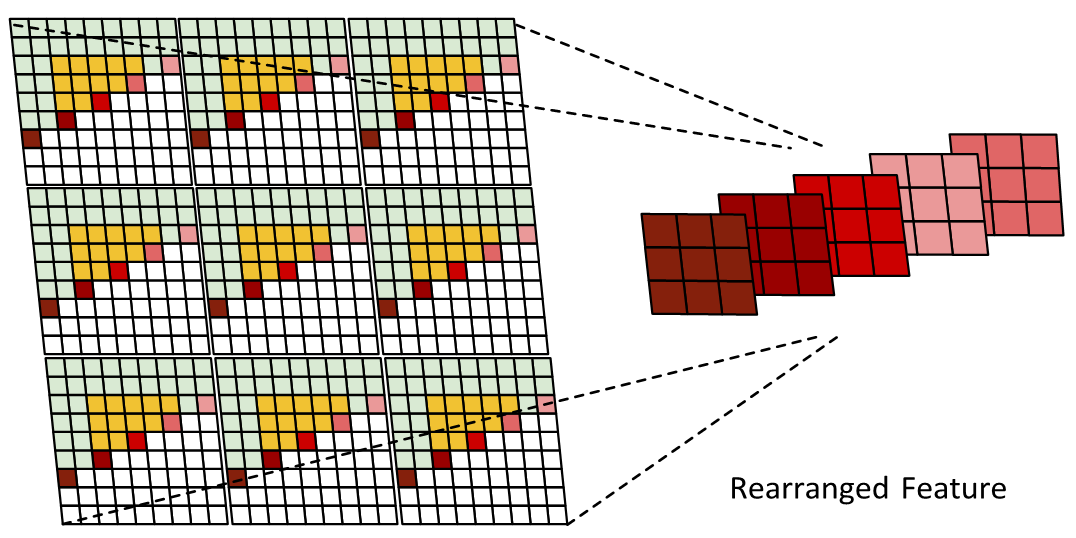}
 \caption{Illustration of the proposed hierarchical context modeling strategy. The context is aggregated from a small neighborhood of adjacent patches, and the context for all patches is computed in parallel. }
\label{fig:extract_context}
\end{figure}

\section{Formulation and Modeling \label{sec:foundations}}
\subsection{Hierarchical Autoregressive Factorization}
\label{sssec:group_formation}

Autoregressive (AR) models factorize the joint probability of data, such as image pixels $\mathbf{x}$, into a product of conditional probabilities:
\begin{equation}
p(\mathbf{x}) = \prod_{i=1}^{N} p(x_i | \mathbf{x}_{<i}),
\label{eq:ar_general}
\end{equation}
where $\mathbf{x}_{<i}$ denotes all previous pixels in a chosen raster-scan order. Classic AR models like PixelCNN~\cite{van2016conditional} adhere to this sequential generation, which is prohibitively slow for large images.

Our goal is to \textbf{rethink} this formulation to enable parallelism at two levels: within patches and across patches.

\subsubsection{Group-wise Parallelism within Patches}
\label{sssec:group_formation_within_patches}

First, to accelerate inference, we divide the image into non-overlapping $P \times P$ patches. Within each patch, we break the strict serial dependency of Eq.~\eqref{eq:ar_general} by organizing pixels into an ordered sequence of $M$ groups, $\mathcal{G}_1, \ldots, \mathcal{G}_{M}$. The factorization is now defined at the group level. The probability for a group $\mathcal{G}_s$, conditioned on all prior groups $\mathcal{G}_{<s}$, is modeled as a product of conditionally independent pixel probabilities:
\begin{equation}
p(\mathcal{G}_s | \mathcal{G}_{<s}) = \prod_{x \in \mathcal{G}_s} p(x | \mathcal{G}_{<s}).
\label{eq:group_ar_intra}
\end{equation}
This formulation is key: all pixels $x$ \textit{within} the same group $\mathcal{G}_s$ are predicted in parallel, as their probabilities depend only on \textit{previous groups} $\mathcal{G}_{<s}$, not on each other.

We define these groups using a parameterized parallel scan order, similar to schemes in~\cite{bai2024deep, li2024callic}. For a pixel $x_{r,c}$ at row $r$ and column $c$ within a patch ($r, c \in \{0, \ldots, P-1\}$), we assign it to a group indexed by $s$:
\begin{equation}
s(r,c) = c + r \cdot \delta.
\label{eq:scan_function}
\end{equation}
Here, $\delta \ge 0$ is a stride parameter that controls the scan pattern. All pixels with the same index $s$ form the group $\mathcal{G}_s$. The groups are then processed in ascending order of $s$, from $s=1$ to $s_{\max} = (1+\delta) P - \delta$. This scan order defines the autoregressive dependency and ensures that every pixel adheres to the same pattern of local contextual dependencies. The parameter $\delta$ directly controls the number of sequential steps required, significantly impacting coding speed. For example, a columnar scan ($\delta=0$) requires only $P$ steps, while a sequential scan ($\delta=P$) requires $P^2$ steps, making the former much faster.

\subsubsection{Hierarchical Parallelism across Patches}

The factorization in Eq.~\eqref{eq:group_ar_intra} only models dependencies \textit{within} a patch. This is a significant limitation, as it overlooks the potential statistical redundancies that exist \textit{between} patches.

To capture this inter-patch context, we introduce a \textbf{hierarchical} dependency. We modify the conditional probability of Eq.~\eqref{eq:group_ar_intra} to also depend on an external context $\mathcal{C}_s$, which summarizes information from other patches. The full joint probability distribution for the entire image is now factorized hierarchically:
\begin{equation}
p(\mathbf{x}) = \prod_{\text{patches}} \left[ \prod_{s=0}^{s_{\max}} \prod_{x \in \mathcal{G}_s} p(x | \mathcal{G}_{<s}, \mathcal{C}_s) \right].
\label{eq:hierarchical_ar_full}
\end{equation}
In this model, $\mathcal{G}_{<s}$ represents the intra-patch context (previous groups in the same patch), while $\mathcal{C}_s$ represents the inter-patch context (information from neighboring patches).

Modeling the global context $\mathcal{C}_s$ is challenging. Inspired by local modeling approaches~\cite{zhang2021out}, we approximate it by aggregating information from a small neighborhood of adjacent patches. For any given pixel, its context is aggregated from features at the same relative position within these neighboring patches. This operation can be implemented efficiently: as shown in Fig.~\ref{fig:extract_context}, by stacking the patch features into a 3D tensor, the aggregation becomes equivalent to a standard 2D convolution across the patch dimension. This allows the context $\mathcal{C}_s$ for all patches to be computed in parallel. Formally, we model the context for group $\mathcal{G}_s$ in a patch $p$ as:
\begin{equation}
\mathcal{C}_{s}(p) = \text{Conv}(\{\mathcal{F}_s(p')\}_{p' \in \mathcal{N}(p)}),
\label{eq:spm_context}
\end{equation}
where $\mathcal{F}_s(p')$ is the feature representation for group $s$ in a neighboring patch $p'$, and $\mathcal{N}(p)$ defines the local neighborhood of patches around $p$.

\subsection{Instance Adaptation from an MDL Perspective}
\label{ssec:mdl}

The Minimum Description Length (MDL) principle provides a formal framework for model selection grounded in information theory~\cite{grunwald2005minimum, barron1998minimum}. It posits that the optimal model for a dataset is the one that permits the greatest compression of the data, which includes the description of the model itself. A prominent formulation is the two-part (or two-stage) MDL principle. This approach partitions the total codelength into two parts: first, the codelength $L(q)$ required to describe a chosen model $q$ from a model class $\mathcal{Q}$, and second, the codelength $L_q(\mathbf{x})$ required to encode the data $\mathbf{x}$ using that model. The objective is to select the model $q$ that minimizes the total codelength:
\begin{equation}
\mathcal{L}_{\text{2-part}}(\mathbf{x}) = \min_{q \in \mathcal{Q}} \left[ L(q) + L_q(\mathbf{x}) \right],
\label{eq:mdl_2part}
\end{equation}
where $L(q) = -\log w(q)$ is the codelength for $q$ under a prior $w(q)$, reflecting model complexity, and $L_q(\mathbf{x}) = -\log q(\mathbf{x})$ is the codelength for $\mathbf{x}$ under $q$.

In the standard paradigm of learned lossless compression, a model is pre-trained on a large dataset. At inference time, this model is assumed to be known to both encoder and decoder, effectively treating its description length $L(q)$ as zero. Only the image bitstream, $L_q(\mathbf{x})$, is transmitted. However, this "one-model-fits-all" approach suffers from an \textbf{amortization gap}: a model optimized for a dataset is sub-optimal for any specific test image, which has its own unique statistics.

To bridge this gap, we propose to adapt the model to each individual image, guided by the two-part MDL principle. A naive approach would be to fine-tune the entire model for each image, but transmitting the full updated weights would make $L(q)$ prohibitively large, defeating the purpose of compression.

Instead, we draw inspiration from Parameter-Efficient Transfer Learning (PETL). We keep the large pre-trained model parameters $\bm{\theta}$ fixed and introduce a small set of trainable, incremental weights $\bm{\phi}$. The pre-trained parameters $\bm{\theta}$ are shared knowledge (zero cost), while only the compact weights $\bm{\phi}$ constitute the "model" to be described. The MDL objective is thus reformulated to jointly optimize the bitrate for these incremental weights and the image data:
\begin{equation}
\min_{\bm{\phi}} \left[ L(\bm{\phi}) + \log \frac{1}{q(\mathbf{x}; \bm{\theta}, \bm{\phi})} \right],
\label{eq:loss_function}
\end{equation}
where $L(\bm{\phi}) = -\log w(\bm{\phi})$ is the codelength for the incremental weights and $-\log q(\mathbf{x}; \bm{\theta}, \bm{\phi})$ is the codelength for the image under the adapted model.

The expected codelength of this scheme with respect to the true data distribution $p$ is given by:
\begin{equation}
\label{eq:d2_m_kl}
L(\bm{\phi}) + \mathbb{E}_{p}\left[-\log q(\mathbf{x}; \btheta, \bm{\phi})\right] = L(\bm{\phi}) + D_{KL}(p\| q) + H(p),
\end{equation}
where $D_{KL}(p\| q)$ is the Kullback-Leibler divergence between the true data distribution $p$ and the adapted model $q$, and $H(p)$ is the entropy of $p$. The expected redundancy $R_n(p)$, or the penalty for not knowing $p$, can be bounded:
\begin{equation}
\label{eq:bound_mdl}
\begin{aligned}
R_n(p) & = E_{p}\left[\mathcal{L}_{\text{2-part}}(\mathbf{x}) - (-\log p(\mathbf{x}))\right] \\
& = E_{p}\left[\min_{\bm{\phi}}\left\{L(\bm{\phi}) + D_{KL}(p\| q(\cdot ; \bm{\theta}, \bm{\phi}))\right\}\right] \\
& \leq \min_{\bm{\phi}} E_{p}\left[L(\bm{\phi}) + D_{KL}(p\| q(\cdot ; \bm{\theta}, \bm{\phi})) \right] \\
& = \min_{\bm{\phi}}\left\{L(\bm{\phi}) + D_{KL}(p \| q(\cdot ; \bm{\theta}, \bm{\phi}))\right\}.
\end{aligned}
\end{equation}
The inequality holds due to Jensen’s inequality. This bound elegantly captures the fundamental trade-off: investing more bits in the model description ($L(\bm{\phi})$) allows for a more complex adaptation, which can reduce the KL divergence ($D_{KL}(p\|q)$) and improve data fidelity. This theoretical formulation underpins our proposed adaptation method, which centers on efficiently optimizing the MDL objective in Eq.~\eqref{eq:loss_function} for each test image.

\section{Proposed Approach} 
\label{sec:method}
\subsection{Hierarchical Parallel Autoregressive ConvNet}
\label{ssec:overall_architecture}

\subsubsection{Convolutional Gating Mechanism \label{sssec:cgm}}
Transformers are celebrated for their outstanding performance~\cite{vaswani2017attention, dosovitskiy2020image}, driven by content-adaptive modeling and long-range dependency capture, but their quadratic complexity hinders high-resolution applications. To efficiently harness content adaptation, we introduce the Convolutional Gating Mechanism (CGM), illustrated in Fig.~\ref{fig:network}(a). Inspired by~\cite{zhang2021out}, CGM distills the adaptive feature modulation of self-attention into an efficient local operation using a depth-wise separable convolution ($k \times k$ kernel) followed by a gating operation (element-wise multiplication). This allows adaptive feature refinement based on local content without the heavy cost of attention.

Mathematically, the general CGM mechanism is:
\begin{equation}
    \begin{gathered}
    \mathbf{A} = \mathrm{DWConv}_{k\times k}(\mathbf{W}_A \mathbf{X}), \\
    \mathbf{V} = \mathbf{W}_V \mathbf{X}, \\
    \mathrm{CGM}(\mathbf{X}) = \sigma(\mathbf{A}) \odot \mathbf{V},
    \end{gathered}
\end{equation}
\label{eq:cgm_general}
where $\mathbf{X}$ is the input, $\mathbf{W}_A, \mathbf{W}_V$ are linear projections, $\sigma$ is swish activation, and $\odot$ is element-wise product.

When causal modeling is required, as in the autoregressive components of HPAC, we employ a specific variant: Masked CGM (MCG). This applies a binary mask $\mathcal{M}_{\delta} \in \{0,1\}^{k \times k}$ (derived from the scan order $\delta$, Section~\ref{sssec:group_formation}) element-wise to the kernel of the depth-wise convolution:
\begin{equation}
\begin{gathered}
    \mathbf{A}_{M} = \mathrm{MDWConv}_{k\times k}(\mathbf{W}_A \mathbf{X}, \mathcal{M}_{\delta}), \\
    \mathrm{MCG}(\mathbf{X}) = \sigma(\mathbf{A}_{M}) \odot \mathbf{V},
\end{gathered}
\label{eq:mcg_masked}
\end{equation}
where $\mathrm{MDWConv}$ denotes the masked depth-wise convolution. Unless specified otherwise (e.g., within the SPM module), CGM operations within HPAC utilize this masked MCG variant to ensure causality.

\subsubsection{Overall Model Architecture}
The Hierarchical Parallel Autoregressive ConvNet (HPAC) model implements our hierarchical parallel modeling principle using CGM blocks. Drawing inspiration from the MetaFormer paradigm~\cite{Yu2022Metaformer, liu2022convnet}, HPAC modularizes network design but replaces generic token mixers with our efficient CGM variants to ensure strict causality where needed, while capturing dependencies at both intra-patch and inter-patch scales.

The overall data flow (Fig.~\ref{fig:network}(b)) begins by normalizing the input image $\mathbf{I}$ to $\tilde{\mathbf{I}}$. Following PixelCNN~\cite{van2016conditional}, an initial masked convolution uses a Type B mask ($\mathcal{M}_{B, \delta}$) to prevent self-conditioning, while subsequent blocks use a Type A mask ($\mathcal{M}_{A, \delta}$) allowing self-conditioning. The core structure consists of this initial convolution, followed by $N$ HPAC-Blocks, and a final $1 \times 1$ convolution to output parameters $\mathbf{\Phi}_{\text{lmm}}$ for the logistic mixture model~\cite{salimans2016pixelcnn++}:
\begin{equation}
\begin{gathered}
    \tilde{\mathbf{I}} = \frac{\mathbf{I}}{2^b} \cdot (v_{\mathrm{max}} - v_{\mathrm{min}}) + v_{\mathrm{min}}, \\
    \mathbf{X}_0 = \mathrm{Conv}_{3\times3}(\tilde{\mathbf{I}}, \mathcal{M}_{B, \delta}), \\
    \mathbf{X}_l = \mathrm{HPAC\text{-}Block}_l(\mathbf{X}_{l-1}), \quad l = 1, \ldots, N, \\
    \mathbf{\Phi}_{\text{lmm}} = \mathrm{Conv}_{1\times1}(\mathbf{X}_N).
\end{gathered}
\label{eq:hpac_flow_revised}
\end{equation}

Each HPAC-Block contains three modules in sequence: a Local Context Modulator (LCM), a channel-mixing MLP, and a Spatial Propagation Module (SPM). Each module uses residual connections with LayerNorm (LN) at the input and LayerScale at the output:
\begin{equation}
\begin{gathered}
    \mathbf{X}'_l = \mathrm{LayerScale}(\mathrm{LCM}(\mathrm{LN}(\mathbf{X}_{l-1}), \mathcal{M}_{A, \delta})) + \mathbf{X}_{l-1}, \\
    \mathbf{X}''_l = \mathrm{LayerScale}(\mathrm{MLP}(\mathrm{LN}(\mathbf{X}'_l))) + \mathbf{X}'_l, \\
    \mathbf{X}_l = \mathrm{LayerScale}(\mathrm{SPM}(\mathrm{LN}(\mathbf{X}''_l))) + \mathbf{X}''_l.
\end{gathered}
\label{eq:hpac_block_flow_revised}
\end{equation}
The LCM employs the masked MCG mechanism (Eq.~\ref{eq:mcg_masked}) with a Type A mask to model fine-grained local dependencies within the patch while respecting causality. The MLP performs channel-wise feature refinement using two linear layers with GELU activation. Finally, the SPM captures broader inter-patch context using its "view-and-convolve" strategy (Fig.~\ref{fig:extract_context}). It rearranges the feature map, applies an efficient non-masked CGM (Eq.~\ref{eq:cgm_general}) across the patch dimension for further feature interaction, before reshaping the tensor back. This enriches each feature vector with context propagated from surrounding patches, relying on the rearrangement for causal consistency rather than masking within the SPM's convolution.

\subsubsection{Efficient Inference via Cache-then-Select Inference}
\label{ssec:csi}

While training benefits from full image context, inference for the HPAC codec must proceed in a group-wise causal manner. To improve the efficiency of this autoregressive generation, we introduce Cache-then-Select Inference (CSI), an optimized strategy that minimizes redundant computation in masked convolutions. As shown in Fig.~\ref{fig:network}(b), CSI works by caching previously computed activations and replacing the standard masked convolution with an efficient selective gather-and-multiply primitive. This primitive gathers only the causally relevant input features for each output location and computes the result via matrix multiplication with the corresponding active kernel weights.

The inference process partitions the image into patches, serializing pixels within each patch into groups $\bm{x} = \{\bm{x}_{\mathcal{G}_1}, \dots, \bm{x}_{\mathcal{G}_{M}}\}$ according to a parallel scan order (see Section~\ref{sssec:group_formation}). This defines $M$ sequential autoregressive steps. At each step $s$, the model processes the current group $\bm{x}_{\mathcal{G}_s}$, predicting its probability distribution by leveraging features from all preceding groups $\bm{x}_{\mathcal{G}_{<s}}$, which are maintained in a state cache.

The implementation of CSI involves two main components. First, a state cache, $\mathbf{X}^{\mathrm{(state)}}$, is maintained for each masked convolutional layer to store activations from processed pixel groups, enabling feature reuse across overlapping receptive fields. Second, the efficient gather-and-multiply primitive relies on active kernel weights, $\mathcal{M}' \in \mathbb{R}^{C_{\mathrm{out}} \times C_{\mathrm{in}} \times N_A}$, which are extracted from the full weight tensor $\mathcal{M}_{\delta}$ in a one-time preprocessing step based on the $N_A$ active locations in the causal mask.

To process a group $\bm{x}_{\mathcal{G}_s}$ containing $N_{\mathrm{eff}}$ pixels, the gather-and-multiply primitive first selectively gathers the $N_A$ causal input features for each of the $N_{\mathrm{eff}}$ output locations from the state cache $\mathbf{X}^{\mathrm{(state)}}$. This forms a compact active input tensor $\mathbf{X}' \in \mathbb{R}^{N_{\mathrm{eff}} \times C_{\mathrm{in}} \times N_A}$. The output features $\mathbf{Y} \in \mathbb{R}^{N_{\mathrm{eff}} \times C_{\mathrm{out}}}$ are then computed via a batched matrix multiplication between the active inputs $\mathbf{X}'$ and active weights $\mathbf{W}'$:
\begin{equation}
Y_{n,o} = \sum_{c=1}^{C_{\mathrm{in}}} \sum_{a=1}^{N_A} X'_{n, c, a} \cdot W'_{o, c, a},
\label{eq:selective_conv_rewrite}
\end{equation}
where $n$ indexes output locations, $o$ output channels, $c$ input channels, and $a$ active kernel positions. This process integrates into the layer-by-layer data flow. The activations from previous layers are first written into the current layer's cache $\mathbf{X}^{\mathrm{(state)}}$, serving as inputs for future groups ($\bm{x}_{\mathcal{G}_{j>s}}$). For a masked $k \times k$ convolution on group $\bm{x}_{\mathcal{G}_s}$, the necessary causal inputs are gathered from the state cache map. In contrast, standard $1 \times 1$ convolutions operate directly on the current group's features and do not require this caching and gathering mechanism. For SPM, as formulated in Eq.~\ref{eq:spm_context}, only the same group positions across different patches are required for context propagation, so the causality is still guaranteed.

\begin{figure}[t]
 \centering
 \includegraphics[width=1.\linewidth]{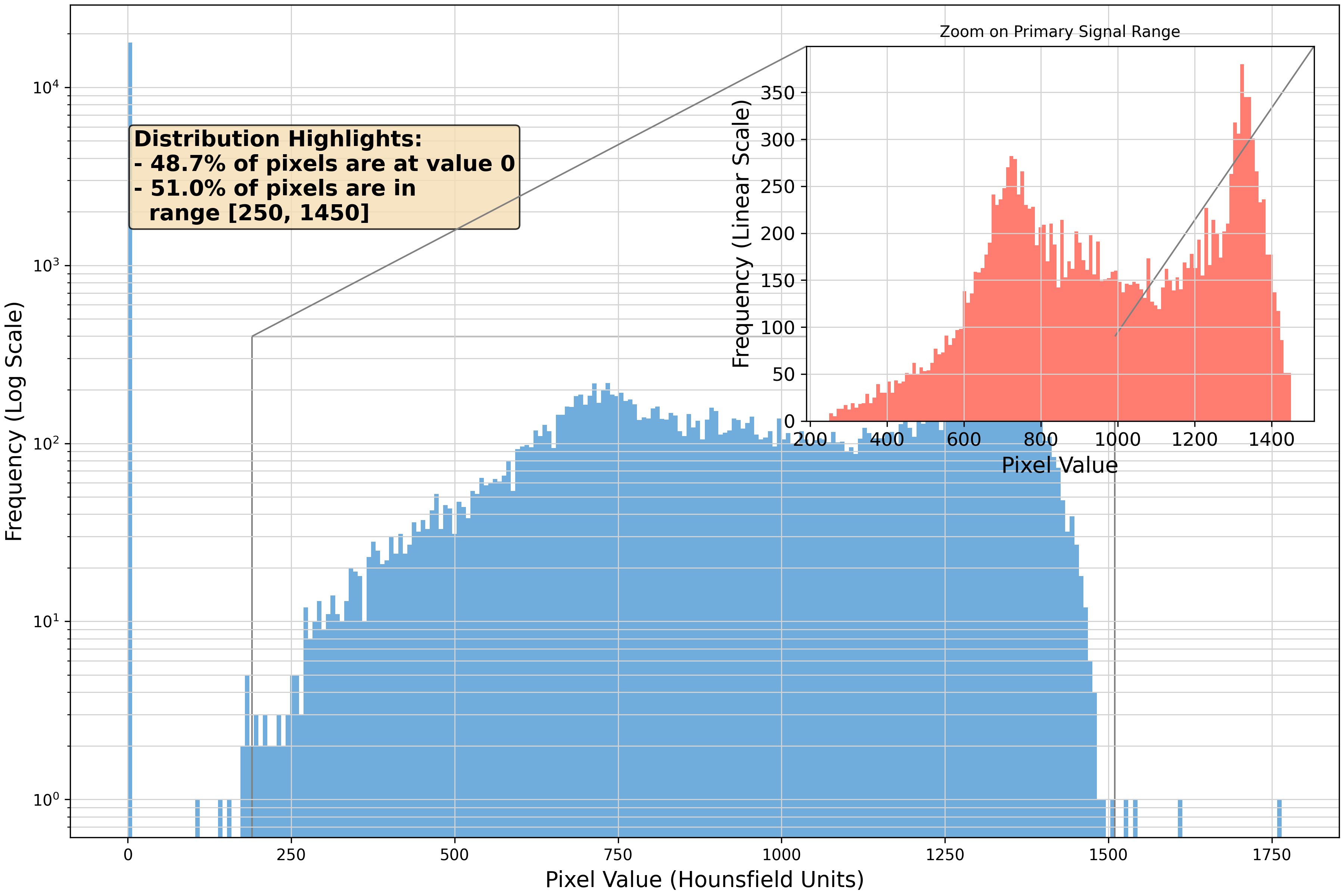}
 \caption{The sparse distribution of pixel values in a sample high bit-depth image from Covid-CT dataset. This distribution is highly skewed, with most pixels concentrated in a narrow range, and only a small fraction of pixels are outliers.
 }
\label{fig:hbd_histogram}
\end{figure}

\subsubsection{Extension for High Bit-depth Images \label{ssec:extending_for_hbd}}
Compressing high bit-depth (HBD) images is challenging due to their wide dynamic range and the common requirement for lossless quality, especially in fields like medical imaging. A typical HBD image contains pixel values concentrated in very narrow regions, with sparse outliers elsewhere that must be perfectly preserved. To quantify this, Fig. \ref{fig:hbd_histogram} illustrates the sparse distribution of pixel values in a sample  high bit-depth medical image. The plot, with a logarithmic y-axis, reveals that the vast majority of pixels are concentrated in specific areas. Specifically, 48.7\% of pixels are at value 0 and another 51.0\% are within the primary signal range of [250, 1450].

This highly skewed distribution poses a key challenge for learning-based lossless codecs. A naive approach would require modeling a probability mass function (PMF) over the entire value space. The PMF table used by entropy coders thus grows exponentially with bit-depth $b$ (size $2^b$), leading to prohibitive memory and computational costs for a distribution that is mostly empty. 

This empirical observation motivates our approach to leverage the inherent compactness of the data distribution. We introduce Adaptive Focus Coding (AFC), an efficient coding-time strategy that avoids instantiating the full $2^b$ PMF table by dynamically "focusing" the probability calculation.
Instead of using a static alphabet, our AFC method dynamically defines a compact coding window for each pixel based on the model's prediction. First, we compute the expected center $\hat{x}_{\text{center}}$ of the predicted Logistic Mixture Model (LMM). This center is calculated as the expected value of the mixture distribution. Given the predicted means $\{\mu_k\}_{k=1}^K$ and mixture weight logits $\{w_k\}_{k=1}^K$ for the $K$ components, $\hat{x}_{\text{center}}$ is the weighted average of the component means:
\begin{equation}
\hat{x}_{\text{center}} = \sum_{k=1}^K \pi_k \mu_k, \quad \text{where} \quad \pi_k = \frac{\exp(w_k)}{\sum_{j=1}^K \exp(w_j)}.
\end{equation}
We then define a dynamic, fixed-size coding window of range $R$ (e.g., $R=1024$) "focused" around this prediction. 
This results in a truncated alphabet $[\hat{x}_L, \hat{x}_U]$, where the bounds are defined by rounding and clamping to the valid pixel range $[0, 2^b-1]$:
\begin{equation}
\begin{gathered}
    \hat{x}_L = \max(0, \lfloor \hat{x}_{\text{center}} - R/2 \rceil),  \\
    \hat{x}_U = \min(2^b-1, \lfloor \hat{x}_{\text{center}} + R/2 \rceil).
\end{gathered}
\end{equation}

The model then computes the PMF only for the discrete values $x$ within this truncated window $[\hat{x}_L, \hat{x}_U]$. Crucially, the probability mass captured within this window is generally less than 1.0. We therefore renormalize this truncated PMF so that its probabilities sum to 1, creating a valid, local probability distribution. This local PMF is then converted to a high-precision CDF table (e.g., 16-bit) for the entropy coder.

To encode a symbol $x$, we first map it to a local index $s = x - \hat{x}_L$. If this index $s$ falls within the local alphabet (i.e., $0 \le s < R'$ where $R' = \hat{x}_U - \hat{x}_L$ is the clamped range), it is encoded directly using the renormalized CDF table. Conversely, if $s$ is an outlier (i.e., $s < 0$ or $s \ge R'$), an escape mechanism is triggered. In this case, we first encode a special sentinel symbol (e.g., $R'$) using the learned CDF, which signals the decoder to switch to a bypass mode. The out-of-range index $s$ is then mapped to a non-negative residual $s_{\text{res}}$ using the bijective function:
\begin{equation}
s_{\text{res}} =
\begin{cases}
 2(s - R') & \text{if } s \geq R' \\
 -2s - 1 & \text{if } s < 0
\end{cases}
\end{equation}
This residual is subsequently encoded using a universal, fixed probability distribution (e.g., a Golomb-Rice code).

This AFC strategy allows the model to leverage the precision of the learned LMM over a compact range, drastically reducing the computational and memory overhead of the entropy coder while still ensuring lossless reconstruction for all HBD values.

\subsection{Progressive Adaptation via SARP-FT} \label{ssec:adaptation}

\subsubsection{Parameter-Efficient Low-Rank Updates \label{ssec:mplora}}

To capture the unique statistical characteristics of each image, we employ a PETL strategy to adapt the pre-trained model weights. Our approach is predicated on the low-rank hypothesis, which posits that the necessary weight adjustments for model adaptation reside in a low-dimensional subspace~\cite{hu2021lora}.

For linear layers, we adopt the Low-Rank Adaptation (LoRA) method. Given a pre-trained weight matrix $\mathbf{W} \in \mathbb{R}^{m \times n}$, the adaptation is modeled as an incremental update $\Delta \mathbf{W}$, which is decomposed into the product of two low-rank matrices. The adapted weight matrix $\mathbf{W'}$ is thus computed as:
\begin{equation}
    \label{eq:lora_linear}
    \mathbf{W'} = \mathbf{W} + \Delta \mathbf{W} = \mathbf{W} + \mathbf{A} \mathbf{B},
\end{equation}
where $\mathbf{A} \in \mathbb{R}^{m \times r}$ and $\mathbf{B} \in \mathbb{R}^{r \times n}$, with the rank $r \ll \min(m, n)$.

To fine-tune the HPAC architecture, we extend this principle from 2D matrices to the 3D tensors of depth-wise convolutional kernels. For a pre-trained kernel $\mathbf{W_{dw}} \in \mathbb{R}^{m \times 1 \times k \times k}$, where $m$ is the channel count and $k$ is the kernel size, we model the update tensor $\Delta \mathbf{W_{dw}}$ using Tucker decomposition~\cite{tucker1966some}. Specifically, the update is structured as the mode-n product of a core identity tensor $\mathbf{I} \in \mathbb{R}^{r_1 \times 1 \times r_2 \times r_3}$ and three low-rank factor matrices: $\mathbf{A} \in \mathbb{R}^{m \times r_1}$, $\mathbf{C} \in \mathbb{R}^{k \times r_2}$, and $\mathbf{D} \in \mathbb{R}^{k \times r_3}$. The update is formulated as:
\begin{equation}
    \Delta \mathbf{W_{dw}} = \mathbf{I} \times_1 \mathbf{A} \times_3 \mathbf{C} \times_4 \mathbf{D},
\end{equation}
where $\times_n$ denotes the tensor-matrix product along mode $n$. The final adapted kernel of depth-wise convolution and its masked version, incorporating the spatial mask $\mathbf{M}$, are:
\begin{equation}
    \begin{gathered}
    \label{eq:lora_conv}
    \mathbf{W'}_{\text{dw}} = \mathbf{W}_{\text{dw}} + \Delta \mathbf{W_{\text{dw}}}, \\
    \mathbf{W'}_{\text{mc}} = \mathbf{M} \odot (\mathbf{W}_{\text{mc}} + \Delta \mathbf{W_{\text{dw}}}),
    \end{gathered}
\end{equation}

A significant advantage of this low-rank formulation is its efficiency. During inference, the incremental weights $\Delta \mathbf{W}$ and $\Delta \mathbf{W}_{\text{dw}}$ can be merged directly into their corresponding pre-trained weights, introducing zero computational overhead. Within our model, we apply these low-rank adaptations to the weight matrices $\mathbf{W}_{\text{A}}$ and $\mathbf{W}_{\text{V}}$ of the convolutional gating mechanism and to the first linear layer $\mathbf{W}_{\text{up}}$ of the MLP. For depth-wise convolution and its masked version, we apply the low-rank adaptation to the weight tensor $\mathbf{W}_{\text{dw}}$ and $\mathbf{W}_{\text{mc}}$ respectively.

\subsubsection{Spatially-Aware Rate-Guided Progressive Fine-tuning} \label{ssec:srpft} 
\begin{figure}[t]
    \centering
    \includegraphics[width=1.\linewidth]{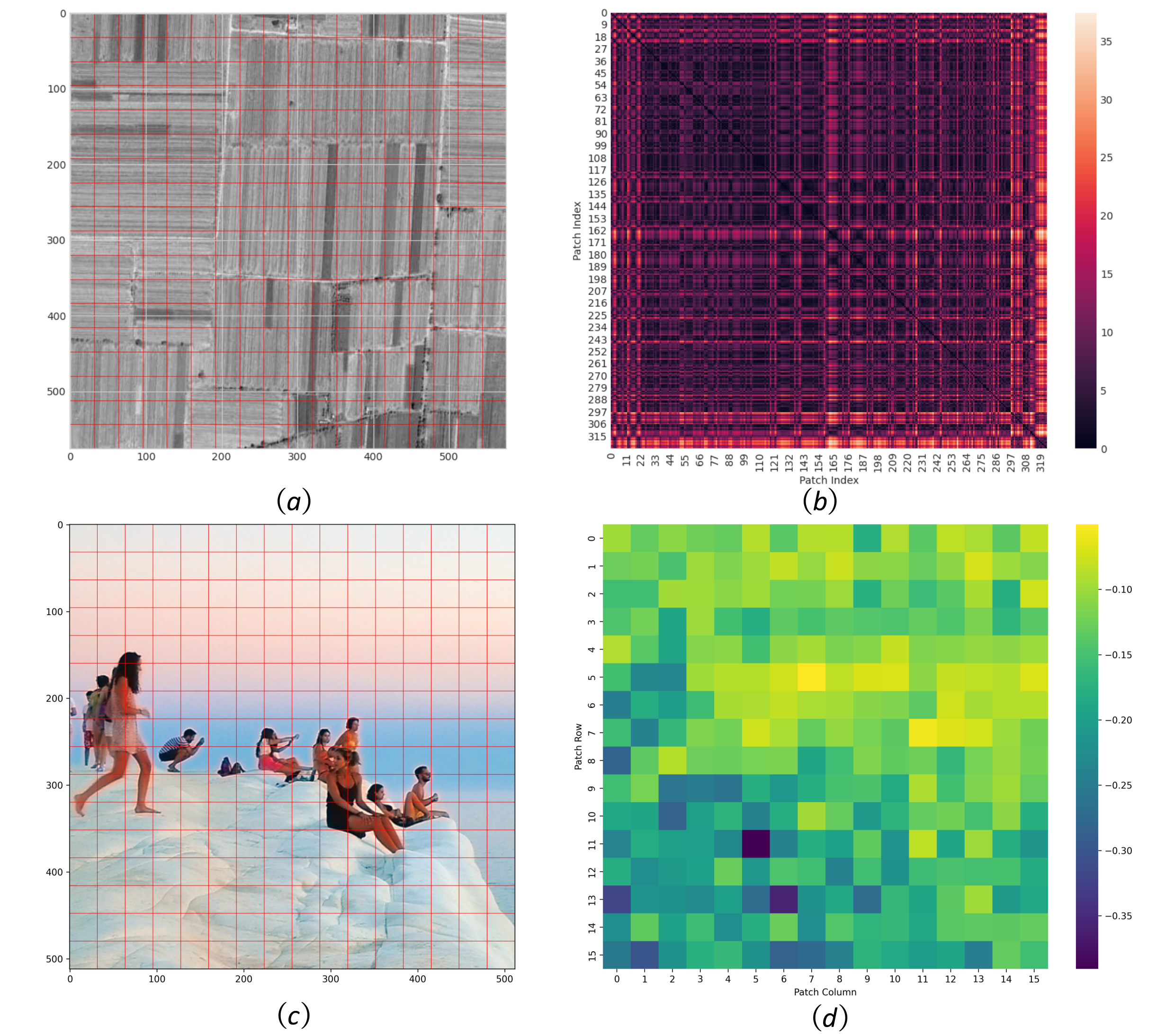}
    \caption{(a) (b) The wassertstein distance between all patch pairs for image \textit{Farmland-49.png} in RS19 dataset, showing the content redundancy across patches. (c) (d) The bpsp gain analysis comparison of fine-tuning on different patches with 10 steps for cropped image \textit{avide-ragusa-716.png} in CLIC.p dataset. The bpsp gain is calculated as the difference in bpsp between the fine-tuned model and the pre-trained model.}
   \label{fig:rpft_motivation}
   \end{figure}

While per-image fine-tuning improves compression, uniformly adapting the model on all image patches is computationally expensive and inefficient. As illustrated in Fig.~\ref{fig:rpft_motivation}, this inefficiency stems from two primary observations. First, significant content redundancy across patches leads to redundant computation. As shown in Fig.~\ref{fig:rpft_motivation}(a), many patch pairs exhibit high distributional similarity, measured by the Wasserstein distance. Second, patches vary in complexity and information density, meaning they contribute unequally to the learning process. Consequently, the performance gains from fine-tuning differ substantially across the image, as observed in Fig.~\ref{fig:rpft_motivation}(b).

An effective strategy, therefore, is to dynamically allocate computational resources to the most informative image regions. We hypothesize that a patch's initial estimated bitrate serves as a reliable proxy for its information content. However, the prior approach of selecting high-bitrate patches individually~\cite{li2024callic} is misaligned with HPAC's architecture. HPAC is a hierarchical autoregressive network designed to capture dependencies among adjacent patches. Selecting a scattered set of high-rate patches violates the assumption of spatial contiguity, depriving the model of the coherent local context required for effective training. 

To overcome these limitations, we introduce Spatially-Aware Rate-Guided Progressive Fine-Tuning (SARP-FT). This strategy is built on the core principles of being rate-guided and progressive: it dynamically allocates computation to the most informative regions (identified by bitrate) and progressively expands this focus from a small starting area to the full image. This directly addresses the inefficiency of uniform adaptation. To resolve the architectural misalignment, the strategy is also made spatially-aware: it identifies a contiguous rectangular region with the highest aggregate information density, rather than scattered patches. This ensures the model adapts on a coherent local context, aligning the fine-tuning process with our model's hierarchical dependency structure for a more effective update.

\begin{algorithm}[t]
    \caption{Lossless Image Compression with SARP-FT \label{alg:compression_process}}
    \begin{algorithmic}[1]
    \Procedure{SARP-FT}{$\bm{x}$, $\btheta$, $\bphi_{\text{init}}$, $T_{\text{adapt}}$, $\eta$}
        \State $\bphi \gets \bphi_{\text{init}}$
        \State Pre-compute Rate Map $M$ and Integral Image $I$
        \For{$t = 1 \to T_{\text{adapt}}$}
            \State $S_t^{*} \gets \text{FindOptimalRegion}(M, I, t)$
            \State $\bm{x}_{S_t^{*}} \gets \text{GetPatches}(\bm{x}, S_t^{*})$
            \State $q(\bm{x}_{S_t^{*}}; \btheta, \bphi) \gets \text{HPAC}(\bm{x}_{S_t^{*}};\btheta, \bphi)$
            \State $\mathcal{L}_{\text{MDL}} \gets \text{Compute Loss using Eq.~\eqref{eq:mdl_opt_loss}}$
            \State $\bphi \gets \bphi - \eta \nabla_{\bphi} \mathcal{L}_{\text{MDL}}$ \Comment{Gradient descent}
        \EndFor
        \State \textbf{return} $\bphi$ \Comment{Return adapted parameters}
    \EndProcedure
    \State
    \Procedure{Encode}{$\mathbf{x},  \btheta, \bphi_{\text{init}}$}
        \State \Comment{Base model $\btheta$, initial adapters $\bphi_{\text{init}}$}
        \State $\bphi_{\text{adapted}} \gets \Call{SARP-FT}{\mathbf{x}, \btheta, \bphi_{\text{init}}}$
        \State $\hat{\bphi} \gets \text{Quantize}(\bphi_{\text{adapted}})$
        \State $q(\cdot; \btheta, \hat{\bphi}) \gets \text{HPAC}(\cdot; \btheta, \hat{\bphi}) $ \Comment{Get adapted model $q$}
        \State $b_{\bphi} \gets  \text{AE}(\hat{\bphi})$ \Comment{Code adapter $\hat{\bphi}$}
        \State $b_{\mathbf{x}} \gets \text{AE}(\mathbf{x} \mid q(\cdot; \btheta, \hat{\bphi}))$ \Comment{Code image $\mathbf{x}$}
        \State $b_{\text{stream}} \gets b_{\bphi} \oplus b_{\mathbf{x}}$ \Comment{Combine bitstreams}
        \State \textbf{return} $b_{\text{stream}}$
    \EndProcedure
    \State
    \Procedure{Decode}{$b_{\text{stream}}, \btheta$}
        \State $b_{\bphi}, b_{\mathbf{x}} \gets \text{Parse}(b_{\text{stream}})$
        \State $\hat{\bphi} \gets \text{AD}(b_{\bphi})$ \Comment{Decode adapter $\hat{\bphi}$}
        \State $q(\cdot; \btheta, \hat{\bphi}) \gets \text{HPAC}(\cdot; \btheta, \hat{\bphi})$ \Comment{Reconstruct model $q$}
        \State $\mathbf{x} \gets \text{AD}(b_{\mathbf{x}} \mid q(\cdot; \btheta, \hat{\bphi}))$ \Comment{Decode image $\mathbf{x}$}
        \State \textbf{return} $\mathbf{x}$
    \EndProcedure
    \end{algorithmic}
\end{algorithm}

The SARP-FT procedure begins with a single forward pass to estimate the bitrate for every patch, creating a 2D Rate Map $M \in \mathbb{R}^{H \times W}$ that preserves the spatial layout. To enable efficient searching, we pre-compute an Integral Image (or summed-area table) $I$ from the rate map $M$, which allows the sum of bitrates over any rectangular region to be calculated in constant time. The fine-tuning process commences on a small rectangular region $S^*$ that maximizes this sum. As training progresses, the target search area gradually expands. At each step, a new optimal region $S^*$ of the current target size is identified, and only the patches within it are used for the model update. During the final phase of training (the last $d$ fraction of steps), the region expands to cover the entire image, ensuring the fully adapted model is optimized for all image content.

We formalize the search for the optimal region at training step $t$ as an optimization problem. The fraction of the total area to be used, $\alpha_t$, is determined by a scheduling function $F(t)$. The target region size is $(h_t, w_t)$ such that $h_t \times w_t \approx N \cdot \alpha_t$, where $N$ is the total number of patches. The optimal region $S^*_t$ is found by:
\begin{equation}
S^*_t = \underset{S(i,j,h_t,w_t)}{\arg\max} \sum_{(i',j') \in S} M_{i',j'},
\end{equation}
and the selected data is $\mathbf{x}_{S^*_t}$.
This search is performed efficiently using the pre-computed integral image $I$. The scheduling function $F(t) = \alpha_t$ can be defined as:
\begin{equation}
\begin{gathered}
F(t) = b + (1 - b) \cdot [s(t')]^e, \quad \text{where} \quad t' = \frac{t}{T \cdot (1 - d)}, \\
s(t) =
\begin{cases}
0, & \text{if } t < 0, \\
1, & \text{if } t > 1, \\
t^2 (3 - 2t), & \text{otherwise}.
\end{cases}
\end{gathered}
\end{equation}
Here, $T$ is the total number of training steps. The hyperparameters $b$ (initial area ratio), $d$ (final full-image phase duration), and $e$ (growth rate exponent) control the shape of the regional growth curve.

\begin{table*}[!t]
    \centering
    \caption{Compression performance on natural images dataset of the proposed methods and other codecs in terms of bits per subpixel (bpsp). The best is highlighted in \textbf{bold}, and the second is highlighted using \underline{underline}. We use ``FT'' to denote our SPAR-FT for simplicity. }
    \begin{tabular}{llccccccccc}
    \toprule
      Type & Codec & DIV2K & CLIC.p & Kodak & RS19 & Histo24 & LoveDA24 & Doc24 & Average (Savings) & Params. \tabularnewline
    \midrule
    \multirow{3}{*}{Traditional}
        & JPEG2000~\cite{skodras2001jpeg}       & 3.12  & 2.93  & 3.19  & 2.57  & 3.36  & 3.65& 2.53& 3.05 (0\%) & $-$    \tabularnewline
        & FLIF~\cite{sneyers2016flif}         & 2.91  & 2.72  & 2.90  & 2.18  & 3.23  & 3.50& 2.54 & 2.85 (6.6\%) & $-$   \tabularnewline
        & JPEG-XL~\cite{alakuijala2019jpeg}    & 2.79  & 2.63  & 2.87  & 2.02  & 3.07  & 3.32 & 2.30 & 2.71 (11.1\%) & $-$   \tabularnewline
    \midrule
    \multirow{10}{*}{Pre-trained}
        & L3C~\cite{mentzer2019practical}        & 3.09   &  2.94  & 3.26  & 2.66  & 3.53  & 3.69& 3.26 & 3.20 (-4.9\%) & 5M   \tabularnewline
        & RC~\cite{mentzer2020learning}         & 3.08  &  2.93  & 3.38  & 2.19  & 3.33  & $-$ & $-$ & $-$ & 6.9M     \tabularnewline
        & iVPF~\cite{zhang2021ivpf}           & 2.68   & 2.54  & $-$ & $-$ & $-$ & $-$ & $-$ & $-$ & 59.5M   \tabularnewline
        & PILC~\cite{kang2022pilc}           & 3.41 & $-$ & $-$ & $-$ & $-$ & $-$ & $-$ & $-$ & $-$   \tabularnewline
        & LC-FDNet~\cite{rhee2022lc}          & 2.72   & 2.63  & 2.98  & 2.15  & 3.07  & 3.34 & 2.72 & 2.80 (8.2\%) & 23.7M \tabularnewline
        & DLPR~\cite{bai2024deep}             & 2.55  & \underline{2.38}   & 2.86   & 2.01  & 2.96  & 3.27 & 2.68 & 2.67 (12.5\%) & 22.2M \tabularnewline
        & ArIB-BPS~\cite{Zhang2024ArIBBPS} & 2.55 &2.42 &2.78 & 1.92  & \underline{2.92} & \underline{3.23} & 2.62 & 2.63 (13.8\%) & 146.6M \tabularnewline
        & FNLIC-Prefiter~\cite{zhang2025fitted} & $-$ & $-$ & 3.36 & $-$ & $-$ & $-$ & $-$ & $-$ & $-$ \tabularnewline
        \arrayrulecolor{gray}
        \cmidrule(lr){2-11}
        \arrayrulecolor{black}
         & \cellcolor{gray!15}HPAC-Fast (Ours)               & \cellcolor{gray!15}2.68 & \cellcolor{gray!15}2.53  & \cellcolor{gray!15}2.95 & \cellcolor{gray!15}1.94  & \cellcolor{gray!15}3.08 & \cellcolor{gray!15}3.41 & \cellcolor{gray!15}2.56 & \cellcolor{gray!15}2.74 (10.2\%) & \cellcolor{gray!15}\textbf{249K}\tabularnewline
         & \cellcolor{gray!15}HPAC (Ours)                &   \cellcolor{gray!15}\underline{2.52} & \cellcolor{gray!15}\underline{2.38}  & \cellcolor{gray!15}\underline{2.73}   & \cellcolor{gray!15}\underline{1.65}  & \cellcolor{gray!15}2.96  & \cellcolor{gray!15}3.27 & \cellcolor{gray!15}\underline{2.46} & \cellcolor{gray!15}\underline{2.57} (\gain{\underline{15.7\%}}) & \cellcolor{gray!15}677K \tabularnewline
    \midrule
    \multirow{3}{*}{Over-fit}
        & FNLIC~\cite{zhang2025fitted} & 2.75 & $-$ & 2.88 & $-$ & $-$ & 3.30 & 2.22 & $-$ & \underline{614K} \tabularnewline
         \arrayrulecolor{gray}
         \cmidrule(lr){2-11}
         \arrayrulecolor{black}
        & \cellcolor{gray!15}HPAC-Fast-FT (Ours)         & \cellcolor{gray!15}2.63 & \cellcolor{gray!15}2.48 & \cellcolor{gray!15}2.74 & \cellcolor{gray!15}1.85 & \cellcolor{gray!15}2.95 & \cellcolor{gray!15}3.28 & \cellcolor{gray!15}2.18 & \cellcolor{gray!15}2.59 (15.1\%) & \cellcolor{gray!15}\textbf{249K}\tabularnewline 
        & \cellcolor{gray!15}HPAC-FT (Ours)         & \cellcolor{gray!15}\textbf{2.47}  & \cellcolor{gray!15}\textbf{2.34} & \cellcolor{gray!15}\textbf{2.52}  & \cellcolor{gray!15}\textbf{1.63} & \cellcolor{gray!15}\textbf{2.81} & \cellcolor{gray!15}\textbf{3.20} & \cellcolor{gray!15}\textbf{1.94} & \cellcolor{gray!15}\textbf{2.42} (\red{\textbf{20.7\%}}) & \cellcolor{gray!15}677K \tabularnewline
        \bottomrule
    \end{tabular}
    \label{tb:results_ll}
\end{table*}

\begin{table}[!t]
    \centering
    \caption{Compression performance (bpsp) on high bit-depth datasets. Best results are in \textbf{bold}, second best are \underline{underlined}. We use ``FT'' to denote our SPAR-FT for simplicity. ID and OOD denote In-domain and Out-of-domain datasets respectively.}
    \begin{tabular}{llccc}
    \toprule
    \multirow{2}{*}{Type} & \multirow{2}{*}{Codec} & \multicolumn{1}{c}{ID} & \multicolumn{1}{c}{OOD} & \multirow{2}{*}{Params.} \\
    & & Covid & Chaos & \\
    \midrule
    \multirow{3}{*}{Traditional}
        & JPEG2000~\cite{skodras2001jpeg} & 6.42 & 5.31 & - \\
        & FLIF~\cite{sneyers2016flif} & 4.85 & 5.21 & - \\
        & JPEG-XL~\cite{alakuijala2019jpeg} & 4.65 & 4.99 & - \\
    \midrule
    \multirow{3}{*}{Learned}
        & BD-LVIC~\cite{wang2024learning} & 4.43 & 5.01 & 5.6M \\
        & \cellcolor{gray!15}HPAC (Ours) & \cellcolor{gray!15}\underline{4.27} & \cellcolor{gray!15}\underline{4.96} & \cellcolor{gray!15}670K \\
        & \cellcolor{gray!15}HPAC-FT (Ours) & \cellcolor{gray!15}\textbf{4.27} & \cellcolor{gray!15}\textbf{4.74} & \cellcolor{gray!15}670K \\
    \bottomrule
    \end{tabular}
    \label{tb:results_ll}
\end{table}

\subsubsection{MDL-Principled Optimization and Overall Process}
The overall compression process is summarized in Alg. \ref{alg:compression_process}. Let $\btheta$ denote the fixed, pre-trained parameters of the HPAC model, and $\bphi$ represent the tunable adapter parameters (i.e., "incremental weights").

The \textsc{Encode} procedure begins by adapting the model to the specific test image $\mathbf{x}$. This is achieved by calling the \textsc{SARP-FT} procedure to optimize $\bphi$ as an image-specific prompt, while $\btheta$ remains frozen. This optimization is guided by the Minimum Description Length (MDL) principle, which aims to jointly minimize the bitrate required for the adapter parameters $\bphi$ and the image data $\mathbf{x}$.

During the \textsc{SARP-FT} optimization loop, we employ a mixed quantization approach~\cite{tsubota2023universal} to obtain differentiable estimates of the final quantized bitrates. For each training step $t$ on a region $\mathbf{x}_{S^*_t}$: The weights used for the forward pass (i.e., data likelihood) are quantized using a Straight-Through Estimator (STE)~\cite{theis2016lossy}: $\hphi = \mathrm{sg}(\lfloor\frac{\bphi}{w} \rceil w - \bphi) + \bphi$, where $w < 1$ is the quantization step size and $\mathrm{sg}$ is the stop-gradient operation.
The weights used for parameter rate estimation are simulated by adding uniform noise: $\tphi = \bphi + U(-\frac{w}{2}, \frac{w}{2})$.
The total loss function $\mathcal{L}_{\text{MDL}}$ is the sum of the data bitrate and the parameter bitrate:
\begin{equation}
    \label{eq:mdl_opt_loss}
    \mathcal{L}_{\text{MDL}} = \frac{1}{N_{\text{total}}} \left( \underbrace{-\log q(\mathbf{x}_{S^*_t};\btheta, \hphi) }_{\text{Image Bits}} + \underbrace{-\log p_s(\tphi)}_{\text{Parameter Bits}} \right),
\end{equation}
where $N_{\text{total}}$ is the total number of pixels in the image, $q(\cdot)$ is the likelihood from the HPAC model, and $p_s(\tphi)$ is the parameter prior modeled as a static logistic distribution with zero mean and a constant scale $s$.

After the \textsc{SARP-FT} procedure converges, the final optimized parameters $\bphi_{\text{adapted}}$ are quantized to $\hat{\bphi}$. As shown in the \textsc{Encode} procedure, the final probability model $q(\cdot; \btheta, \hat{\bphi})$ is determined. Finally, the total bitstream $b_{\text{stream}}$ is generated by entropy coding the quantized parameters $\hat{\bphi}$ (yielding $b_{\bphi}$) and the image $\mathbf{x}$ (yielding $b_{\mathbf{x}}$).

As detailed in the \textsc{Decode} procedure, the decoder first parses the bitstream, decodes the adapter weights $\hat{\bphi}$, reconstructs the identical probability model $q(\cdot; \btheta, \hat{\bphi})$, and then sequentially decodes the image pixels $\mathbf{x}$.

\begin{table}[t!]
    \centering
    \setlength{\tabcolsep}{3.5pt} 
    \caption{Coding speed analysis on Kodak. Bpsp denotes bits per sub-pixel. We use ``FT'' to denote our SPAR-FT for simplicity. CCI denotes Cache-then-Crop Inference method in our conference paper~\cite{li2024callic}.}
    \begin{tabular}{llccc}
    \toprule
    Type & Codec & Bpsp & Params. & Latency (s) \\
    \midrule
    \multirow{7}{*}{Pre-trained} 
        & LC-FDNet~\cite{rhee2022lc} & 2.98 & 23.7M & 1.75/1.71 \\
        & DLPR~\cite{bai2024deep} & 2.86 & 22.2M & 1.20/1.42 \\
        & ArIB-BPS~\cite{Zhang2024ArIBBPS} & 2.78 & 146.6M & 11.34/10.28 \\
        & \cellcolor{gray!15}HPAC (Ours) & \cellcolor{gray!15}2.73 & \cellcolor{gray!15}677K & \cellcolor{gray!15}0.89/0.90 \\
        & \textit{-w/ CCI}~\cite{li2024callic} & $-$ & $-$ & \textit{1.01/1.03} \\
        & \textit{-w/o cache inference} & $-$ & $-$ & \textit{10.34/11.07} \\
        & \cellcolor{gray!15}HPAC-Fast (Ours) & \cellcolor{gray!15}2.95 & \cellcolor{gray!15}249K & \cellcolor{gray!15}0.39/0.40 \\
    \midrule
    \multirow{4}{*}{Over-fit} 
        & FNLIC~\cite{zhang2025fitted} & 2.88 & 614K & 6249/0.42 \\
        & \cellcolor{gray!15}HPAC-FT (Ours) & \cellcolor{gray!15}2.52 & \cellcolor{gray!15}677K & \cellcolor{gray!15}8.45/0.94 \\
        & \textit{-w/o SAPR-FT (Full-image)} & \textit{2.54} & $-$ & \textit{13.91/0.94} \\
        & \cellcolor{gray!15}HPAC-Fast-FT (Ours) & \cellcolor{gray!15}2.74 & \cellcolor{gray!15}249K & \cellcolor{gray!15}4.87/0.41 \\
    \bottomrule
    \end{tabular}
    \label{tb:results_time}
\end{table}

\section{Experiments \label{sec:experiments}}

\subsection{Experimental Setup \label{ssec:experimental_setup}}
\subsubsection{Datasets \label{sssec:datasets}}
We evaluate our method on six high-resolution image datasets, selected to cover a diverse range of image types and complexities:
\begin{itemize}
    \item \textit{Kodak}~\cite{kodak}: 24 uncompressed $768 \times 512$ color images, a standard benchmark for natural image compression, 24 images.
    \item \textit{RS19}~\cite{Xia2010WHURS19}: 19-category high-res satellite images for remote sensing, 190 images.
    \item \textit{Histo24}~\cite{bai2024deep}: 24 high-res histopathology images with complex textures, 24 images.
    \item \textit{DIV2K}~\cite{Agustsson_2017_CVPR_Workshops}: 100 diverse, high-quality images for model robustness evaluation.
    \item \textit{CLIC.p}~\cite{clic}: CLIC 2020 professional validation set of high-res photos, 41 images.
    \item \textit{LoveDA24}~\cite{zhang2025fitted}: Remote sensing images, using the validation set from~\cite{zhang2025fitted}, 24 images.
    \item \textit{Doc24}~\cite{zhang2025fitted}: Historical document images from ICDAR’24 MapText Competition, using the validation set from~\cite{zhang2025fitted}, 24 images.
\end{itemize}

Also, additional datasets are used to evaluate our high bit-depth (HBD) images compression performance.
\begin{itemize}
    \item \textit{Covid-CT}~\cite{wang2024learning}: Pulmonary CT volumes. CT-2 is used for training; for testing, we select the highest-entropy slice from each of 24 CT-3 volumes.
    \item \textit{Chaos-CT}~\cite{wang2024learning}: Abdominal CT volumes from 12 patients, from which we select the highest-entropy slices from each of 12 CT volumes for testing.
\end{itemize}

We benchmark our method against traditional lossless codecs, including JPEG2000~\cite{skodras2001jpeg}, FLIF~\cite{sneyers2016flif}, and JPEG-XL~\cite{alakuijala2019jpeg}, as well as open-source learned methods, including the VAE-based L3C~\cite{mentzer2019practical}, the Flow-based iVPF~\cite{zhang2021ivpf}, and hybrid methods such as RC~\cite{mentzer2020learning}, LC-FDNet~\cite{rhee2022lc}, DLPR~\cite{bai2024deep}, ArIB-BPS~\cite{Zhang2024ArIBBPS}, and FNLIC~\cite{zhang2025fitted}. For HBD images compression, we use BD-LVIC~\cite{wang2024learning} trained on Covid-CT training set as strong learned baseline. Compression performance is measured in bits per subpixel (bpsp).
\subsubsection{Pre-training Configuration \label{sssec:pre_training_configuration}}

Our proposed HPAC model is configured with a depth of $N=3$, a hidden dimension of $128$, an MLP expansion ratio of $4$, and uses a $3 \times 3$ kernel for the initial embedding and $7 \times 7$ for subsequent convolutions. The output is modeled by a logistic mixture with $K=5$ components. We use $32 \times 32$ patches and a parallel degree of $\delta=2$, resulting in $94$ autoregressive steps. For a faster variant, HPAC-Fast, we reduce the depth to $N=2$, the mixture components to $K=3$, the patch size to $16 \times 16$, and set the parallel degree to $\delta=1$, hidden dimension to $96$ while keeping the convolution kernel size unchanged.

For 8-bit image datasets, the models were pre-trained on a composite dataset of $9 \times 10^4$ images sourced from DIV2K~\cite{Agustsson_2017_CVPR_Workshops}, FLICKR 2K~\cite{Lim_2017_CVPR_Workshops}, and ImageNet~\cite{deng2009imagenet}.
For HBD images compression, the models were pre-trained on Covid-CT training set. Chaos-CT testing set is used as out-of-domain (OOD) dataset for testing our method's generalization ability.
During training, images were randomly cropped into $128 \times 128$ patches. We trained the models for $2$ million steps using the Adam optimizer with a minibatch size of $8$. A cosine annealing schedule was employed for the learning rate, with a peak value of $1 \times 10^{-3}$.

\subsubsection{Per-Image Adaptation Configuration \label{sssec:per_image_adaptation_configuration}}

For the per-image fine-tuning stage, optimization is performed for a maximum of $T=50$ steps with a learning rate of $1 \times 10^{-2}$ to balance compression performance and adaptation speed. The standard deviation of the logistic prior for the incremental weights was set to $s=0.05$, and the uniform quantization width was set to $w=0.05$. The hyperparameters for our proposed SARP-FT were configured by default to $b=0.2$, $d=0.1$, and $e=1$. The rank for the low-rank decomposition of the incremental weights was set to $r=8$. 

\begin{table*}[t]
    \centering
    \caption{Analysis of network architecture of HPAC. We perform an ablation study on network depth, model channels, masked convolution kernel size, and patch size. The baseline model is highlighted in \textbf{bold}. For performance metrics, `↓` indicates that lower is better.}
    \begin{tabular}{lccccccccc}
        \toprule
        \multirow{2}{*}{Study on} & \multirow{2}{*}{Depth} & \multirow{2}{*}{Channels} & \multirow{2}{*}{Kernel Size} & \multirow{2}{*}{Patch Size $P$} & \multirow{2}{*}{Parallel Degree $\delta$} & \multicolumn{2}{c}{Coding Latency (s)} & \multirow{2}{*}{Params.} & \multirow{2}{*}{Bpsp (↓)} \\
        \cmidrule(lr){7-8}
        & & & & & & Encoding & Decoding & & \\
        \midrule
        \multirow{3}{*}{Network Depth} & 1 & \multirow{3}{*}{128} & \multirow{3}{*}{7} & \multirow{3}{*}{32} & \multirow{3}{*}{2} & 0.60 & 0.62 & 198K & 2.88 \\
        & \textbf{3} & & & & & \textbf{0.89} & \textbf{0.90} & \textbf{677K} & \textbf{2.73} \\
        & 5 & & & & & 1.21 & 1.23 & 1.2M & 2.71 \\
        \midrule
        \multirow{3}{*}{Model Channels} & \multirow{3}{*}{3} & 64 & \multirow{3}{*}{7} & \multirow{3}{*}{32} & \multirow{3}{*}{2} & 0.83 & 0.85 & 179K & 2.79 \\
        & & \textbf{128} & & & & \textbf{0.89} & \textbf{0.90} & \textbf{677K} & \textbf{2.73} \\
        & & 192 & & & & 0.94 & 0.95 & 1.5M & 2.71 \\
        \midrule
        \multirow{4}{*}{Kernel Size} & \multirow{4}{*}{3} & \multirow{4}{*}{128} & 3 & \multirow{4}{*}{32} & \multirow{4}{*}{3} & 0.84 & 0.85 & 662K & 2.76 \\
        & & & 5 & & & 0.87 & 0.88 & 668K & 2.75 \\
        & & & \textbf{7} & & & \textbf{0.89} & \textbf{0.90} & \textbf{677K} & \textbf{2.73} \\
        & & & 9 & & & 0.95 & 0.98 & 689K & 2.73 \\
        \midrule
        \multirow{4}{*}{Patch Size} & \multirow{4}{*}{3} & \multirow{4}{*}{128} & \multirow{4}{*}{7} & 8 & \multirow{4}{*}{2} &  0.58 & 0.53 & 677K & 2.81 \\
        & & & & 16 &  & 0.64 & 0.61 & 677K & 2.77 \\
        & & & & \textbf{32} & & \textbf{0.89} & \textbf{0.90} & \textbf{677K} & \textbf{2.73} \\
        & & & & 64 & & 1.41 & 1.45 & 677K & 2.72 \\
        \midrule
        \multirow{4}{*}{Parallel Degree} & \multirow{4}{*}{3} & \multirow{4}{*}{128} & \multirow{4}{*}{7} & \multirow{4}{*}{32} & 1 & 0.74 & 0.75 & 677K & 2.83 \\
        & & & & & \textbf{2} & \textbf{0.89} & \textbf{0.90} & \textbf{677K} & \textbf{2.73} \\
        & & & & & 3 &1.02 &1.08  & 677K & 2.73 \\
        & & & & & 4 & 1.26 & 1.32 & 677K & 2.72 \\
        \bottomrule
    \end{tabular}
    \label{tab:ab_network_optimized_no_macs}
\end{table*}

\begin{figure*}[t]
    \centering
    \includegraphics[width=\linewidth]{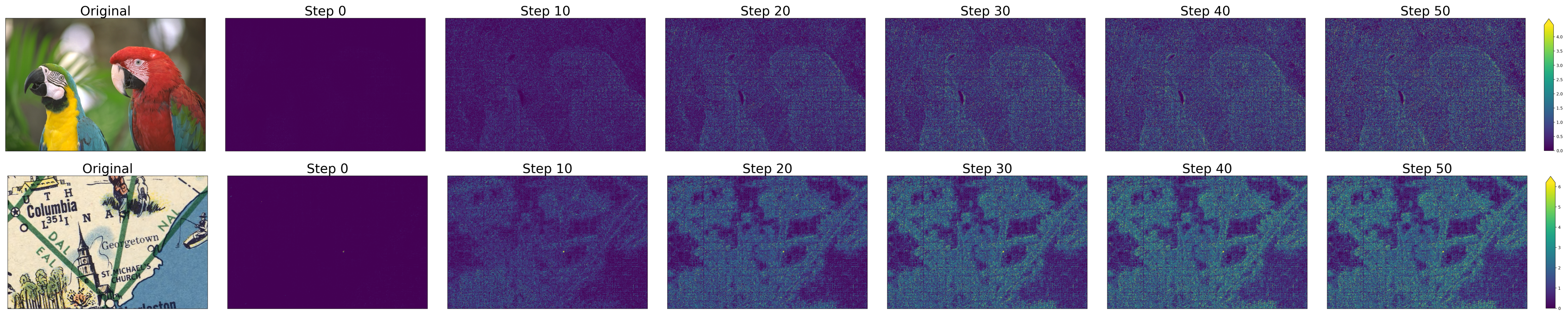}
    \caption{The bpsp savings of SARP-FT over pre-trained HPAC on Kodak and Doc24 datasets. The first row shows \textit{kodim23.png} in Kodak dataset. The second row shows \textit{24.png} in Doc24 dataset. 
    }
   \label{fig:sarp_gain}
   \end{figure*}

\subsection{Experimental Results \label{ssec:experimental_results}}

\subsubsection{Compression Performance}
As summarized in Tab.~\ref{tb:results_ll}, our pre-trained model, HPAC, achieves SOTA performance, leading on five datasets (DIV2K, CLIC.p, Kodak, RS19, Doc24) and matching ArIB-BPS on Histo24 and LoveDA24. Overall, HPAC averages 15.7\% bpsp savings over JPEG-2000 and 1.5\% over the previous SOTA, ArIB-BPS. Crucially, HPAC achieves this with only 677K parameters (0.5\% of ArIB-BPS), as shown in Tab.~\ref{tb:results_time}. Our lightweight variant, HPAC-Fast (249K), also surpasses prior work like LC-FDNet (25.1M) while using just 1\% of its parameters. These results validate the efficiency of our hierarchical autoregressive architecture.

When integrating SARP-FT for instance adaptation, HPAC-FT consistently surpasses all other methods across all datasets. It achieves 20.7\% bpsp savings over JPEG-2000 and 8.0\% over ArIB-BPS. The adaptation impact correlates with the train-test distribution gap: it yields a 20\% bpsp saving on the highly OOD Doc24 dataset, while gains are modest for in-distribution data like CLIC.p. Notably, HPAC-Fast-FT also outperforms adaptive methods like FNLIC while using 60\% fewer parameters.

In HBD compression, HPAC surpasses the BD-LVIC baseline by 3.6\% bpsp on the in-domain Covid-CT dataset. On the OOD Chaos-CT dataset, SARP-FT adds another 4.4\% saving over the pre-trained HPAC, highlighting its effectiveness where standard models falter.

Fig.~\ref{fig:sarp_gain} visualizes the progressive bpsp savings over 50 fine-tuning steps on Kodak and Doc24 images. Starting from zero savings at Step 0, SARP-FT accumulates increasing gains, especially in complex regions. The effect is notably stronger on Doc24 (second row), aligning with the quantitative improvements in Tab.~\ref{tb:results_ll}.

\subsubsection{Coding Speed and Complexity \label{sssec:coding_speed_and_complexity}}
Coding speed and complexity are analyzed in Tab.~\ref{tb:results_time} (Kodak dataset, single NVIDIA RTX 4090D).

For pre-trained models, HPAC's 0.89s encoding latency is significantly faster than SOTA methods like ArIB-BPS (11.34s). This speed is enabled by our cache inference mechanism; removing it increases latency 11-fold (to 10.34s). Our CSI also saves 10\% coding time over the CCI method~\cite{li2024callic} by eliminating redundant computations. HPAC-Fast offers an even lighter solution (249K params) with a 0.39s encoding time, ranking it among the fastest learned methods.

\begin{figure}[t]
 \centering
 \includegraphics[width=0.9\linewidth]{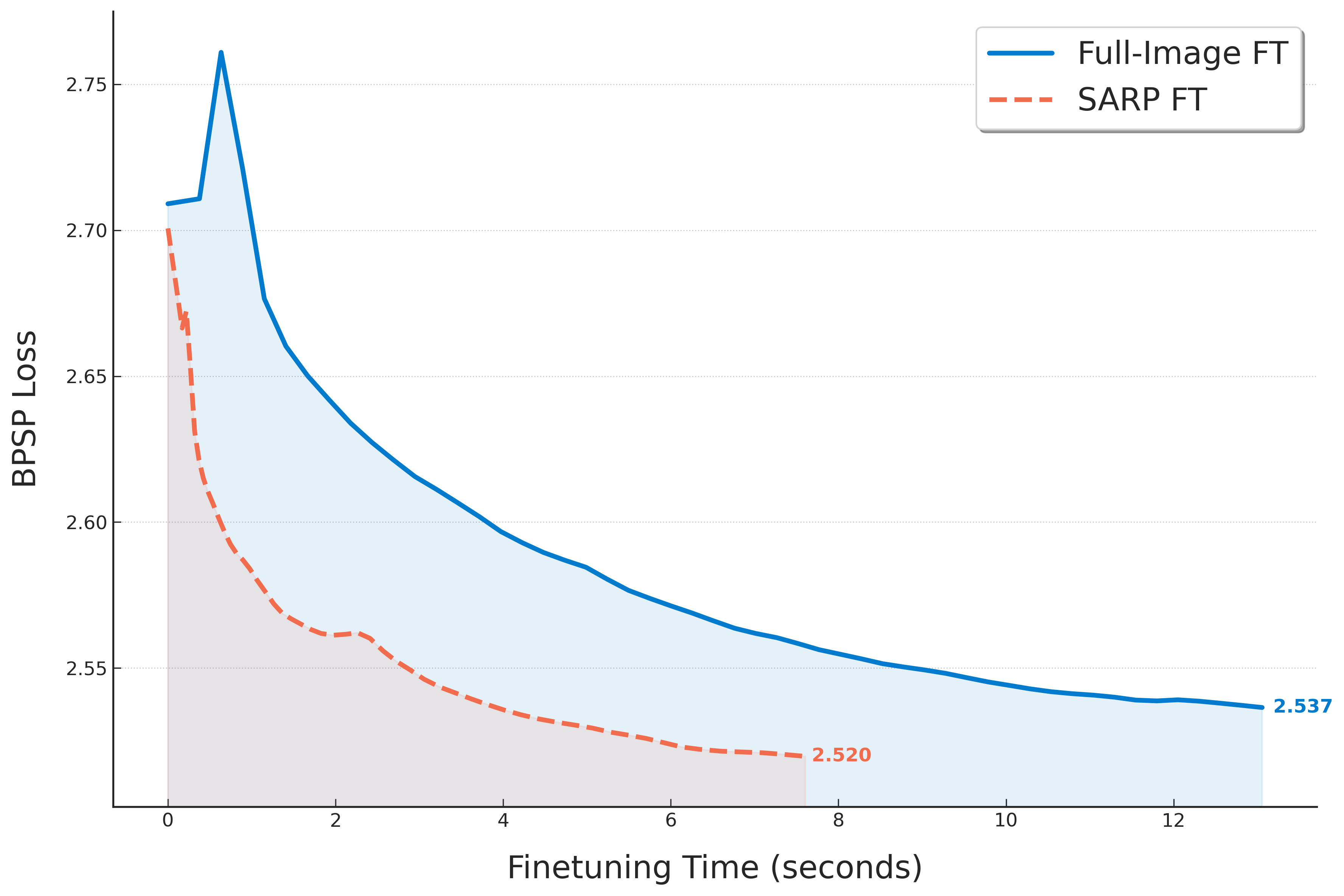}
 \caption{Fine-tuning bpsp loss and time comparison between full-image (Full-Image FT) and our proposed progressive fine-tuning (SARP-FT) on Kodak dataset.}
\label{fig:exp_steps}
\end{figure}

For instance-adaptive methods, SARP-FT demonstrates significant practical advantages. HPAC's lightweight nature is a key enabler, reducing the forward/backward pass burden. Consequently, HPAC-FT achieves SOTA compression (Tab.~\ref{tb:results_ll}) in only 8.45 seconds—a drastic improvement over FNLIC (6249s, or $> 1.5$ hours). Our SARP-FT strategy is also 41.7\% faster than naive full-image fine-tuning (13.05s) and achieves superior compression (2.52 vs. 2.54 bpsp). 
This speedup enables faster convergence (Fig.~\ref{fig:exp_steps}), highlighting a practical performance-time trade-off. For instance, 10 fine-tuning steps yield 2.59 bpsp in only 0.74 seconds, securing a large portion of the compression gains at a fraction of the full adaptation cost.

\subsection{Ablation Studies and Analysis \label{subsec:ablation_studies}}

\subsubsection{Network Architecture Study \label{sssec:network_architecture_study}}
We conduct a comprehensive ablation study to analyze the impact of key hyperparameters on the performance of HPAC, with the results detailed in Tab.~\ref{tab:ab_network_optimized_no_macs}. This analysis covers network depth, model channels, masked convolution kernel size, patch size $P$, and parallel degree $\delta$.

\textbf{Network Depth, Channels, and Kernel Size.} We first examine the trade-off between model complexity and compression performance. As shown in Tab.~\ref{tab:ab_network_optimized_no_macs}, increasing the network depth, channel width, and kernel size generally leads to better compression, indicated by a lower bpsp. For instance, expanding the depth from 1 to 3 reduces the bpsp from 2.88 to 2.73. Similarly, increasing the channels from 64 to 128 lowers the bpsp from 2.79 to 2.73. However, these gains diminish as the model size continues to grow; a further increase in depth to 5 or channels to 192 yields only marginal improvements (0.02 bpsp) while substantially increasing the number of parameters and computational latency. A larger kernel size also enhances performance, but the improvement becomes less pronounced beyond a kernel size of 7. Considering these trade-offs, we select a depth of 3, 128 channels, and a kernel size of 7 for our baseline model.

\textbf{Patch Size.} The choice of patch size $P$ influences the model's ability to capture spatial dependencies. The results show a clear trend: larger patch sizes lead to better compression performance, reducing the bpsp from 2.81 ($P=8$) to 2.72 ($P=64$). This is because larger patches provide a wider receptive field, enabling the model to leverage longer-range correlations. However, this benefit comes at the cost of significantly increased encoding and decoding times. We choose $P=32$ as it offers a competitive compression rate while maintaining a manageable latency.

\textbf{Parallel Degree.} The parallel degree $\delta$ is another critical factor. Increasing $\delta$ from 1 to 2 results in a substantial bpsp reduction from 2.83 to 2.73. However, further increasing $\delta$ to 3 or 4 yields negligible compression gains while progressively increasing the coding latency. Therefore, we set $\delta=2$ as the optimal value for our configuration. These results demonstrate that our group-wise autoregressive factorization is effective, sacrificing negligible compression performance for a significant and controllable speed-up.

\begin{table}[t]
\centering
\caption{Comparison between our Hierarchical Parallel Autoregressive ConvNet (HPAC) and the Neighborhood Attention Transformer (NAT)~\cite{hassani2023neighborhood} on the Kodak dataset. }
\begin{tabular}{lccccc}
\toprule
Method & Bpsp & Params. & Enc. Time & Dec. Time \\
\midrule
HPAC (Ours) & 2.73 & 677K & 0.89s & 0.90s \\
LCM-SPM-MLP & 2.74 & $-$ & $-$ & $-$ \\
SPM-LCM-MLP & 2.76 & $-$ & $-$ & $-$ \\
w/o SPM & 2.79 & 575K & 0.83s & 0.84s \\
NAT & 2.78 & 767K & 7.12s & 7.51s \\
\bottomrule
\end{tabular}
\label{tb:compare_arch_components}
\end{table}

\begin{table}[t!]
  \centering
  \caption{Comparison of bpsp (bits per sub-pixel) and codec time (seconds) and peak memory across different PMF lengths ($R$) on Covid-CT~\cite{wang2024learning} dataset.}
  \label{tab:pmf_length_analysis}
  \begin{tabular}{lccccc}
    \toprule
    $R$ & 256 & 512 & \textbf{1024} & 2048 & 4096 \\
    \midrule
    bpsp& 5.34 & 4.36 & 4.27 & 4.24 & 4.23 \\
    Encode Time (s) & 0.59 & 0.63 & 0.70 & 0.90 & 1.45 \\
    Decode Time (s) & 0.59 & 0.71 & 0.78 & 0.97 & 1.47 \\
    Peak Memory (GB) & 0.69 & 1.19 & 1.93 & 3.28 & 6.47 \\
    \bottomrule
\end{tabular}
\end{table}

\begin{figure}[t]
    \centering
    \includegraphics[width=0.95\linewidth]{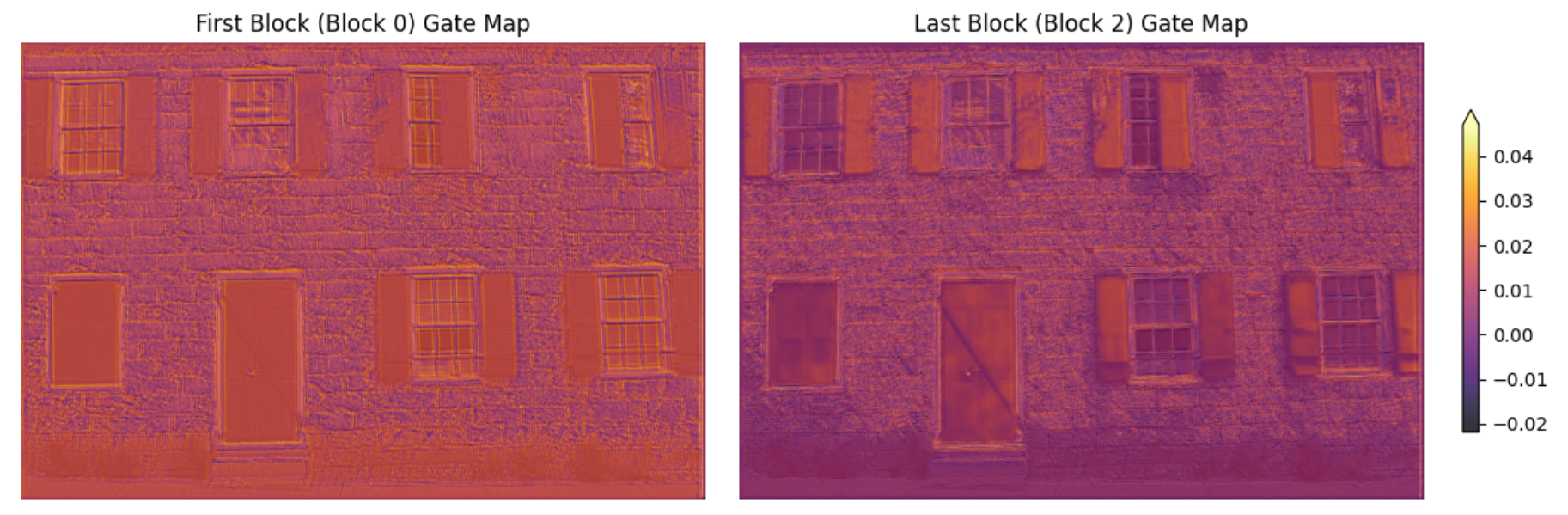}
    \caption{Visualization of the HPAC block activations at different layers on \textit{kodim01.png} in Kodak dataset.}
   \label{fig:hpca_vis}
   \end{figure}

\subsubsection{Visualization of Convolutional Gating Mechanism}
To understand HPAC's internal workings, we visualize the mean gate activations ($\sigma(A)$) from the CGM (Section~\ref{sssec:cgm}). Fig. \ref{fig:hpca_vis} reveals a clear progressive specialization: shallow layers (Block 0) show high, uniform activations, acting as general feature gatherers. In contrast, deep layers (Block 2) become highly sparse and selective, focusing high activations on complex, high-entropy regions (e.g., textures, edges) while suppressing simpler, flat areas. This visually confirms our CGM is content-adaptive and that HPAC progressively refines its focus to efficiently modulate complex spatial dependencies validating our architecture's efficiency and design principles.

\subsubsection{Effectiveness of Masked Gated Convolution \label{sssec:effectiveness_of_masked_gated_convolution}}
We compare our Masked Gated Convolution (MGC) with the Neighborhood Attention Transformer (NAT)~\cite{hassani2023neighborhood}, a local self-attention variant. As shown in Tab.~\ref{tb:compare_arch_components} and visually supported by the analysis in Fig.~\ref{fig:hpca_vis}, MGC achieves better compression (2.73 vs 2.78 bpsp) with 12\% fewer parameters (677K vs. 767K). More importantly, MGC is over 8x faster in both encoding and decoding, thanks to its lightweight convolution operations compared to attention computations. These combined results demonstrate MGC's superior efficiency-performance trade-off.

\subsubsection{Ablation Study on HPAC Block Structure \label{sssec:effectiveness_of_hierarchical_modeling}}
We analyze the design choices within the HPAC block using Tab.~\ref{tb:compare_arch_components}, focusing on module order and the contribution of the SPM. First, comparing different arrangements confirms our default sequence (LCM-MLP-SPM) yields the best compression (2.73 bpsp), slightly outperforming alternatives like LCM-SPM-MLP and SPM-LCM-MLP (both 2.76 bpsp). Second, we assess the impact of inter-patch modeling by removing the SPM modules (`w/o SPM`). This variant achieves 2.79 bpsp. Incorporating the SPM significantly improves the compression rate to 2.73 bpsp, demonstrating the clear benefit of hierarchical modeling across patches. This gain comes with a modest increase in parameters (from 575K to 677K) but a negligible change in latency.

\begin{table}[t!]
    \centering
    \caption{Ablation study on Low-rank Adaptation (LoRA) for the HPAC-FT model on the Kodak dataset.}
    \label{tb:ablation_study}
    \setlength{\tabcolsep}{5pt}
    \begin{tabular}{llcc}
    \toprule
    Component & Setting & Bpsp $\downarrow$ & Latency (s) \\
    \midrule
    \multirow{4}{*}{Rank ($r$)} 
        & 1  & 2.58 & 8.41 \\
        & 4  & 2.54 & 8.41 \\
        & 8  & 2.52 & 8.45 \\
        & 16 & 2.59 & 8.47 \\
    \midrule
    \multirow{3}{*}{Position} 
        & LCM\&SPM & 2.60 & 7.87 \\
        & MLP       & 2.56 & 7.53 \\
        & Both & 2.52 & 8.45 \\
    \bottomrule
    \end{tabular}
    \end{table}

\begin{figure}[t]
 \centering
 \includegraphics[width=0.95\linewidth]{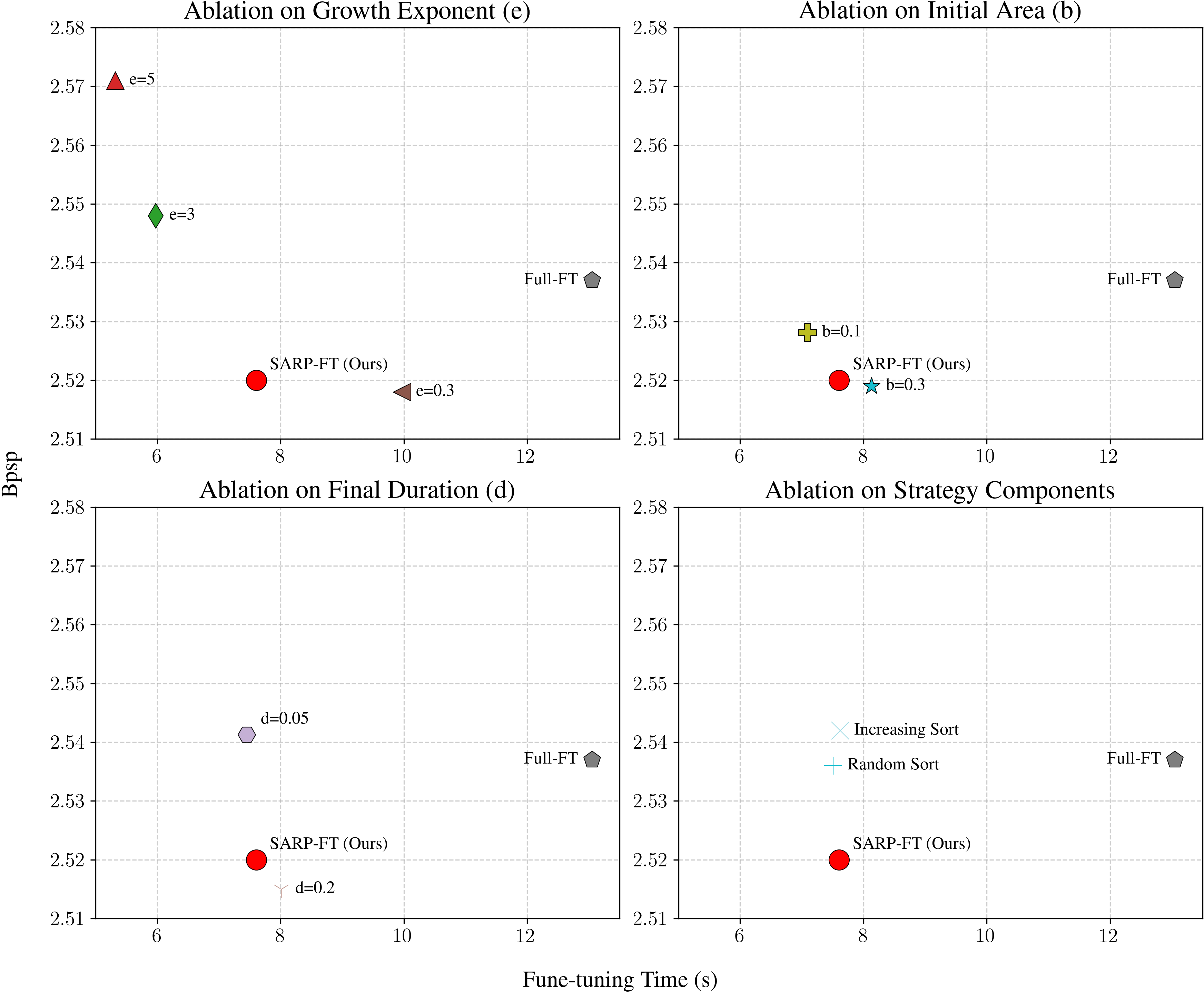}
 \caption{Ablation study of SARP-FT scheduling parameters and sorting strategies on the Kodak dataset.}
\label{fig:ab_RPFT}
\end{figure}

\subsubsection{Ablation Study on Adaptive Focus Coding \label{sssec:ablation_study_on_afc}}
Our proposed AFC enables dynamic adjustment of PMF length based on compression requirements. To understand its impact, we conduct experiments with different PMF lengths ($R$) and analyze their effects on compression performance. The results presented in Tab.~\ref{tab:pmf_length_analysis} reveal that increasing the PMF length from 256 to 4096 leads to improved compression efficiency, reducing the bpsp from 5.34 to 4.23. However, the marginal benefits diminish at larger lengths. This improvement comes with increased computational costs—both encoding and decoding latencies approximately double when scaling from 256 to 4096, while peak memory consumption rises substantially from 0.69GB to 6.47GB. After careful consideration of these trade-offs, we select $R=1024$ as the default PMF length, as it achieves an optimal balance between compression ratio and resource utilization.

\subsubsection{Ablation Study on Low-rank Adaptation \label{sssec:ablation_study_on_how_to_introduce_incremental_parameters}}
We investigate how the type, rank, and position of incremental parameters affect compression performance. The results are shown in Tab.~\ref{tb:ablation_study}. 
Increasing the rank from 1 to 8 generally improves compression, but rank 16 hurts performance (2.59 bpsp). This demonstrates the MDL trade-off between saving data bits and the overhead of transmitting the larger adapter, as indicated in Eq.~\eqref{eq:loss_function}. The position of the parameters also matters, and we find that placing LoRA adapters in all modules (`Both`) achieves the best compression.

\subsubsection{Ablation Study on SARP-FT Configurations \label{sssec:ablation_study_on_srpft_configurations}}
We conduct a detailed ablation study to analyze the impact of different configurations of our SARP-FT strategy, with results presented in Fig.~\ref{fig:ab_RPFT}. Our analysis validates the effectiveness of SARP-FT and the rationale behind our default parameter choices.

First, we compare our default SARP-FT configuration ($b=0.2, e=1, d=0.1$) against a `baseline` that fine-tunes on the entire image. The baseline approach requires 13.05 seconds to achieve a bpsp of 2.537. In contrast, our default SARP-FT is both significantly faster and more effective: it achieves a superior bpsp of 2.520 in only 7.605 seconds. This represents a 41.7\% reduction in time while also improving the compression rate, validating that our progressive, spatially-aware strategy is far more efficient than naive full-image tuning.

Next, we explore the trade-offs defined by the scheduling parameters. As expected, allocating more resources—by increasing the base ratio $b$ (to 0.3), increasing the final duration $d$ (to 0.2), or decreasing the exponent $e$ (to 0.3)—yields even better compression rates (2.519, 2.515, and 2.518 bpsp, respectively) at the cost of longer encoding times. Conversely, accelerating the schedule by increasing $e$ (to 3 or 5) significantly reduces time (to 5.97s and 5.32s) but degrades compression (2.548 and 2.571 bpsp). This confirms that our default parameters ($e=1$) represent a robust sweet spot between speed and performance, while also showing that our framework can be tuned for even faster speeds or higher compression if desired.

The region selection strategy is also critical. The results confirm that our default method of prioritizing high-entropy (i.e., high-rate) patches (2.520 bpsp) is decisively superior to alternative strategies, such as processing patches in order of increasing entropy (2.542 bpsp) or using random sampling (2.536 bpsp), both of which yield significantly worse compression for a similar time cost.

\section{Conclusion \label{sec:conclusion}}
This paper challenged the conventional view that pure autoregressive (AR) models are impractical for lossless image compression. We presented a novel framework built on hierarchical parallelism and progressive adaptation that yields a high-performing, efficient, and adaptive solution.

Our framework introduces the Hierarchical Parallel Autoregressive ConvNet (HPAC), an ultra-lightweight architecture made practical by Cache-then-Select Inference (CSI) for acceleration and Adaptive Focus Coding (AFC) for high bit-depth support. Building on HPAC's efficiency, our Spatially-Aware Rate-Guided Progressive Fine-tuning (SARP-FT) strategy enables rapid, instance-specific adaptation, effectively bridging the amortization gap.
Extensive evaluations confirm our approach sets a new state-of-the-art, significantly outperforming existing methods in compression rate while maintaining competitive coding speeds and a remarkably small parameter footprint. Our work demonstrates that by rethinking model architecture and adaptation, pure AR models can form the basis of a practical, universal solution for lossless image compression.


%





\ifCLASSOPTIONcaptionsoff
  \newpage
\fi



\bibliographystyle{IEEEtran}
\bibliography{IEEEabrv,refs}

\begin{IEEEbiography}
    [{\includegraphics[width=1in,height=1.25in,clip,keepaspectratio]{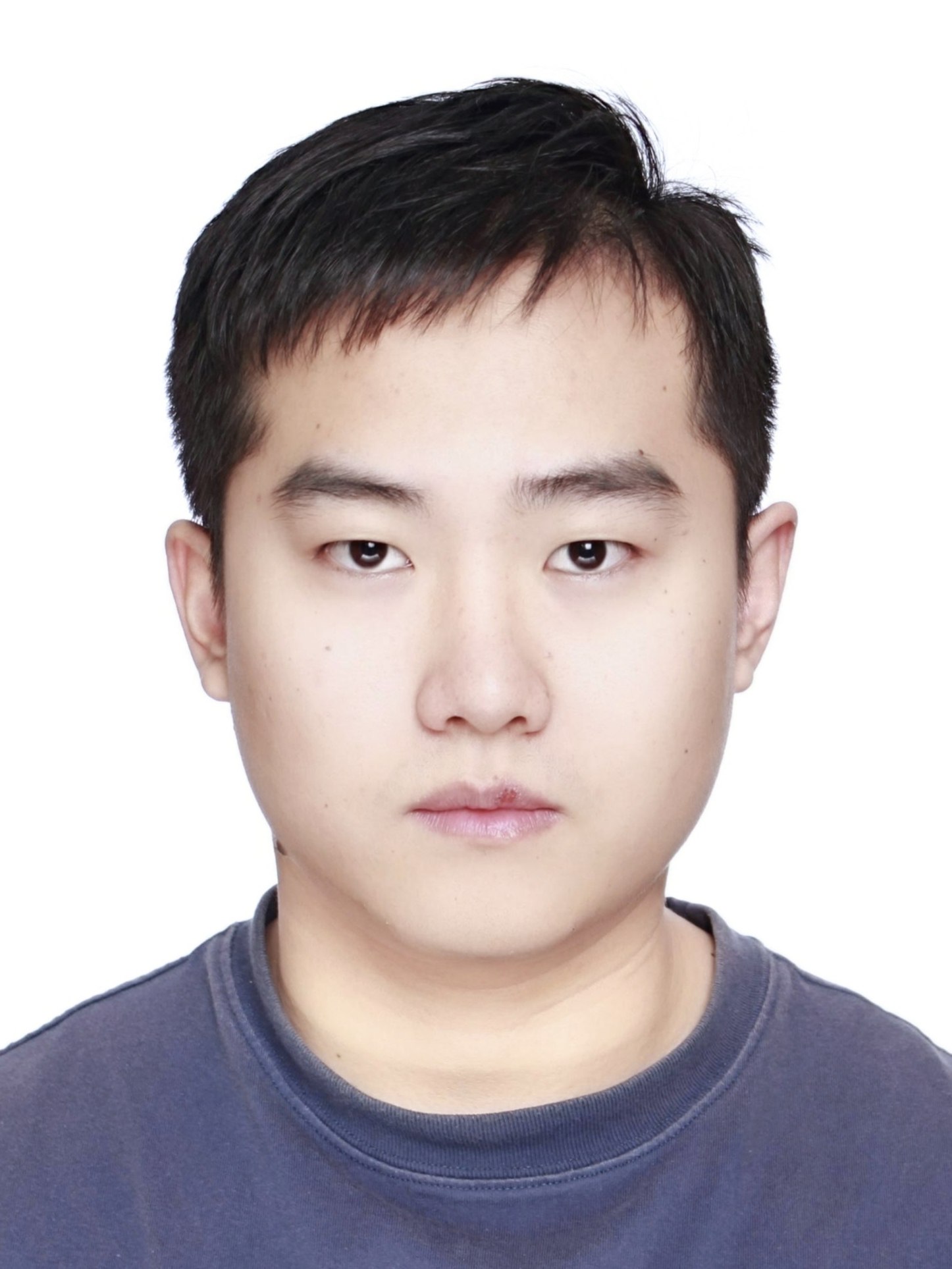}}]{Daxin Li}
    received his B.S. degree in the Faculty of Computation from Harbin Institute of Technology, Harbin, China, in 2021. He is currently pursuing his Ph.D. degree in the Faculty of Computation from Harbin Institute of Technology. His research interests lie primarily in the areas of image and video compression, as well as deep learning applications.
    \end{IEEEbiography}
    \begin{IEEEbiography}[{\includegraphics[width=1in,height=1.25in,clip,keepaspectratio]{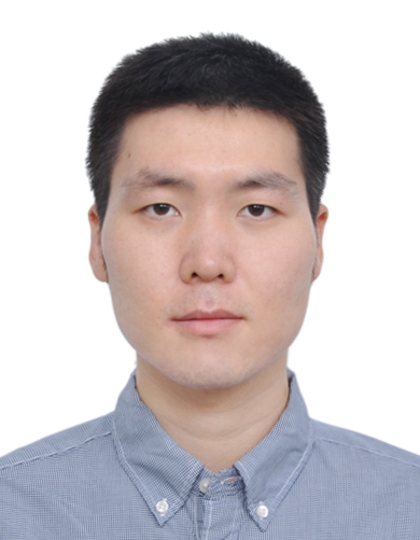}}]{Yuanchao Bai} (Member, IEEE)
    received the B.S. degree in software engineering from Dalian University of Technology, Liaoning, China, in 2013. He received the Ph.D. degree in computer science from Peking University, Beijing, China, in 2020. He was a postdoctoral fellow in Peng Cheng Laboratory, Shenzhen, China, from 2020 to 2022. He is currently an assistant professor  with the School of Computer Science and Technology, Harbin Institute of Technology, Harbin, China.
    His research interests include image/video compression and processing, deep unsupervised learning, and graph signal processing.
    \end{IEEEbiography}
    \begin{IEEEbiography}[{\includegraphics[width=1in,height=1.25in,clip,keepaspectratio]{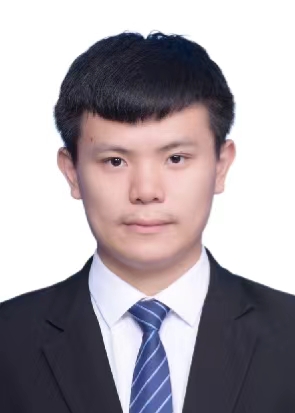}}]{Kai Wang}
    received the B.S. degree in software engineering from Harbin Engineering University, Harbin, China, in 2020 and received the M.S. degree of electronic information in software engineering from Harbin Institute of Technology, Harbin, China, in 2022. He is currently pursuing the docter degree in electronic information in Harbin Institute of Technology, Harbin, China.
    His research interests include image/video compression and deep learning.
    \end{IEEEbiography}

    \begin{IEEEbiography}[{\includegraphics[width=1in,height=1.25in,clip,keepaspectratio]{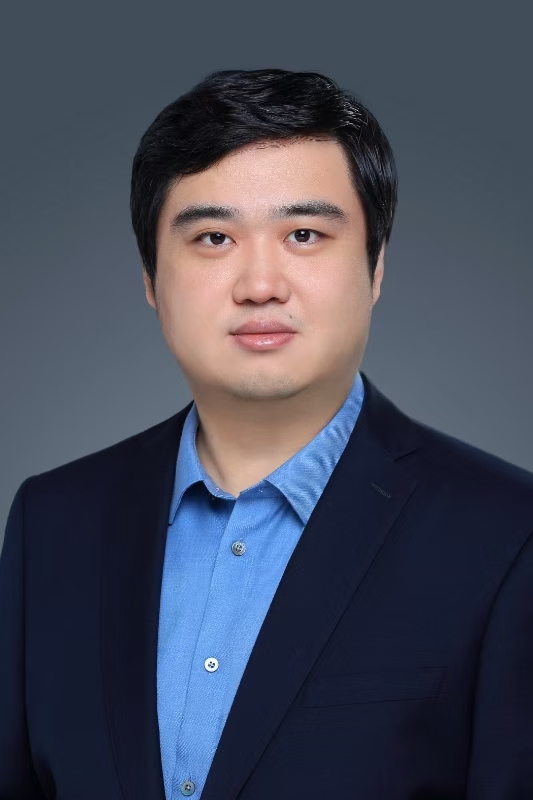}}]{Wenbo Zhao} (Member, IEEE) 
        received the B.S., M.S., and Ph.D. degrees from Harbin Institute of Technology (HIT), Harbin, China, in 2012, 2014, and 2020, respectively. He was also a Post-Doctor at Peng Cheng Laboratory, Shenzhen, China. He is currently an Associate Professor with the School of Computer Science and Technology, Harbin Institute of Technology.  His research interests include mesh denoising, point cloud compression, and deep learning.
        
    \end{IEEEbiography}

    \begin{IEEEbiography}[{\includegraphics[width=1in,height=1.25in,clip,keepaspectratio]{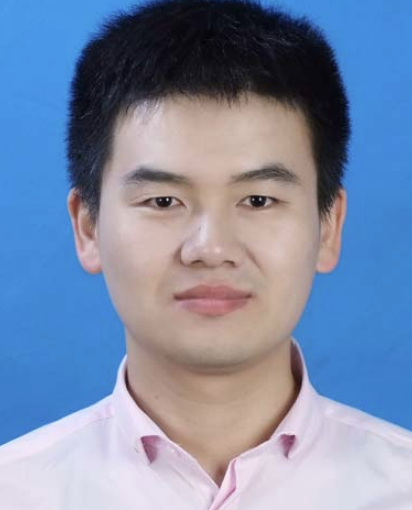}}]{Junjun Jiang} (Senior Member, IEEE) 
    received the B.S. degree in mathematics from Huaqiao University, Quanzhou, China, in 2009, and the Ph.D. degree in computer science from Wuhan University, Wuhan, China, in 2014. From 2015 to 2018, he was an Associate Professor with the School of Computer Science, China University of Geosciences, Wuhan. From 2016 to 2018, he was a Project Researcher with the National Institute of Informatics (NII), Tokyo, Japan. He is currently a Professor with the School of Computer Science and Technology, Harbin Institute of Technology, Harbin, China. His research interests include image processing and computer vision.
    \end{IEEEbiography}
    \begin{IEEEbiography}[{\includegraphics[width=1in,height=1.25in,clip,keepaspectratio]{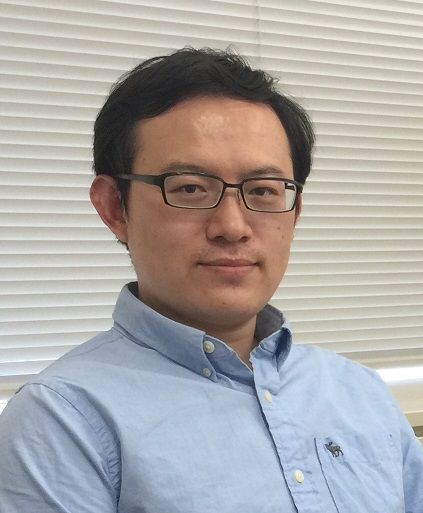}}]{Xianming Liu} (Member, IEEE)
     is a Professor with the School of Computer Science and Technology, Harbin Institute of Technology (HIT), Harbin, China. He received the B.S., M.S., and Ph.D degrees in computer science from HIT, in 2006, 2008 and 2012, respectively. In 2011, he spent half a year at the Department of Electrical and Computer Engineering, McMaster University, Canada, as a visiting student, where he then worked as a post-doctoral fellow from December 2012 to December 2013. He worked as a project researcher at National Institute of Informatics (NII), Tokyo, Japan, from 2014 to 2017. He has published over 200  top-tier international conference and journal publications, including top IEEE journals, such as T-IP, T-CSVT, T-IFS, T-MM, T-GRS; and top conferences, such as CVPR, IJCAI and DCC. He is the receipt of IEEE ICME 2016 Best Student Paper Award.
    \end{IEEEbiography}

\end{document}